\documentclass{article}

\usepackage{microtype}
\usepackage{graphicx}
\usepackage{booktabs} 

\usepackage{hyperref}

\usepackage[accepted]{icml2025}

\usepackage{amsmath}
\usepackage{amssymb}
\usepackage{mathtools}
\usepackage{amsthm}

\usepackage[capitalize,noabbrev]{cleveref}

\theoremstyle{plain}

\theoremstyle{definition}

\theoremstyle{remark}

\usepackage[textsize=tiny]{todonotes}

\usepackage[utf8]{inputenc} 
\usepackage[T1]{fontenc}    
\usepackage{hyperref}       
\usepackage{url}            
\usepackage{booktabs}       
\usepackage{amsfonts}       
\usepackage{nicefrac}       
\usepackage{microtype}      
\usepackage{xcolor}         
\usepackage{subcaption}
\usepackage{wrapfig}
\usepackage[linguistics,edges]{forest}
\usepackage{threeparttable} 
\usepackage{graphicx} 
\usepackage{colortbl}
\usetikzlibrary{decorations.markings}
\usetikzlibrary{shadows}
\usepackage{adjustbox}

\usepackage{hyperref}

\usepackage{tcolorbox}
\usepackage{diagbox}
\definecolor{metablue}{HTML}{0081FB}

\newcommand{\promptbox}[1]{%
    \begin{tcolorbox}[
        colback=white,
        colframe=black,
        arc=4pt,
        boxrule=1pt,
        left=6pt,
        right=6pt,
        top=6pt,
        bottom=6pt
    ]
    \textbf{prompt: } #1
    \end{tcolorbox}
}

\newcommand{\gptoOnebox}[1]{%
    \begin{tcolorbox}[
        colback=orange!5,
        colframe=orange!50!black,
        arc=2pt,
        boxrule=2pt,
        left=8pt,
        right=8pt,
        top=8pt,
        bottom=8pt,
        title={\textcolor{teal!5!white}{\textbf{GPT-o1 Output:}}},
        fonttitle=\small
    ]
    #1
    \end{tcolorbox}
}

\newcommand{\claudesonnetbox}[1]{%
    \begin{tcolorbox}[
        colback=green!5,
        colframe=green!50!black,
        arc=2pt,
        boxrule=2pt,
        left=8pt,
        right=8pt,
        top=8pt,
        bottom=8pt,
        title={\textcolor{green!5!white}{\textbf{Claude 3.5 Sonnet Output:}}},
        fonttitle=\small
    ]
    #1
    \end{tcolorbox}
}

\newcommand{\claudeopusbox}[1]{%
    \begin{tcolorbox}[
        colback=teal!5,
        colframe=teal!60!white,
        arc=2pt,
        boxrule=2pt,
        left=8pt,
        right=8pt,
        top=8pt,
        bottom=8pt,
        title={\textcolor{teal!5!white}{\textbf{Claude 3 Opus Output:}}},
        fonttitle=\small
    ]
    #1
    \end{tcolorbox}
}

\newcommand{\gptfourbox}[1]{%
    \begin{tcolorbox}[
        colback=purple!3,
        colframe=purple!40!black,
        arc=2pt,
        boxrule=2pt,
        left=8pt,
        right=8pt,
        top=8pt,
        bottom=8pt,
        title={\textcolor{purple!5!white}{\textbf{GPT-4o Output:}}},
        fonttitle=\small
    ]
    #1
    \end{tcolorbox}
}

\icmltitlerunning{Visual Graph Arena}

\begin{document}

\twocolumn[
\icmltitle{Visual Graph Arena: Evaluating Visual Conceptualization of \\
Vision and Multimodal Large Language Models}



\icmlsetsymbol{equal}{*}

\begin{icmlauthorlist}
\icmlauthor{Zahra Babaiee}{equal,tuwien}
\icmlauthor{Peyman M. Kiasari}{equal,tuwien}
\icmlauthor{Daniela Rus}{mit}
\icmlauthor{Radu Grosu}{tuwien}

\end{icmlauthorlist}

\icmlaffiliation{tuwien}{CPS, TU Wien, Vienna, Austria}
\icmlaffiliation{mit}{CSAIL, MIT}

\icmlcorrespondingauthor{Peyman Kisari}{peyman.kiasari@tuwien.ac.at}

\icmlkeywords{Machine Learning, ICML}

\vskip 0.3in
]



\printAffiliationsAndNotice{\icmlEqualContribution} 

\begin{abstract}
Recent advancements in multimodal large language models have driven breakthroughs in visual question answering. Yet, a critical gap persists, `conceptualization'—the ability to recognize and reason about the same concept despite variations in visual form, a basic ability of human reasoning. To address this challenge, we introduce the Visual Graph Arena (VGA), a dataset featuring six graph-based tasks designed to evaluate and improve AI systems’ capacity for visual abstraction. VGA uses diverse graph layouts (e.g., Kamada-Kawai vs. planar) to test reasoning independent of visual form. Experiments with state-of-the-art vision models and multimodal LLMs reveal a striking divide: humans achieved near-perfect accuracy across tasks, while models totally failed on isomorphism detection and showed limited success in path/cycle tasks. We further identify behavioral anomalies suggesting pseudo-intelligent pattern matching rather than genuine understanding. These findings underscore fundamental limitations in current AI models for visual understanding. By isolating the challenge of representation-invariant reasoning, the VGA provides a framework to drive progress toward human-like conceptualization in AI visual models. The Visual Graph Arena is available at: \href{https://vga.csail.mit.edu/}{vga.csail.mit.edu}.
\end{abstract}

\section{Introduction}

In recent years, the fields of object recognition~\cite{girshick2015region} and visual question answering~\cite{antol2015vqa, goyal2017making, yang2022empirical} have witnessed remarkable progress, pushing the boundaries of artificial intelligence. Despite these advancements, a gap remains between current AI capabilities and the goal of achieving artificial general intelligence (AGI). One critical aspect of this gap is the ability of AI models to engage in deep reasoning and understand fundamental concepts from visual inputs~\cite{gupta2023visual}, which is essential for reaching human-like cognitive abilities~\cite{han2019visual}. This gap is particularly evident in what we term `conceptualization' - the ability to recognize and reason about concepts across different representations, even when their visual form varies significantly. 


Humans have a long history of developing various methods of visual representation to simplify complex information and enhance understanding. These representations span across diverse domains, such as chemical structures, logic circuits, flowcharts, and statistical data, each with its unique set of visual conventions and abstractions. Figure~\ref{fig:datasamples} showcases the diversity and complexity of visual representations that humans regularly encounter and interpret. In Figure~\ref{fig:datasamples}-a, two different visualizations of the same logic circuit are presented, illustrating how identical functional relationships can be conveyed through varied graphical styles. Similarly, Figure~\ref{fig:datasamples}-b depicts two representations of the same chemical structure, highlighting the challenge of recognizing underlying molecular configurations when presented in different visual forms. Figure~\ref{fig:datasamples}-c features a graph drawn in two distinct layouts, both representing the same underlying data structure and relationships. These examples underscore a crucial aspect of visual reasoning and highlight the core challenge of `conceptualization': the ability to consistently understand different visual representations of the same concept, a task that is typically intuitive for humans but remains challenging for AI.

\newcommand{\gapwidth}{0.49}
\begin{figure*}[ht]
  \centering
  
  \begin{subfigure}{0.32\linewidth}
  \centering
    \includegraphics[width=0.33\linewidth]{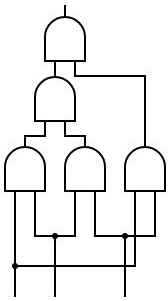}
    \includegraphics[width=0.35\linewidth]{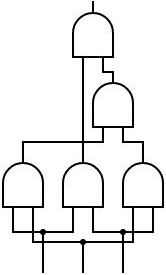}
    \caption{Two representations of a logic circuit.}
  \end{subfigure}
  \hfill
  \begin{subfigure}{0.32\linewidth}
  \includegraphics[width=\gapwidth\linewidth]{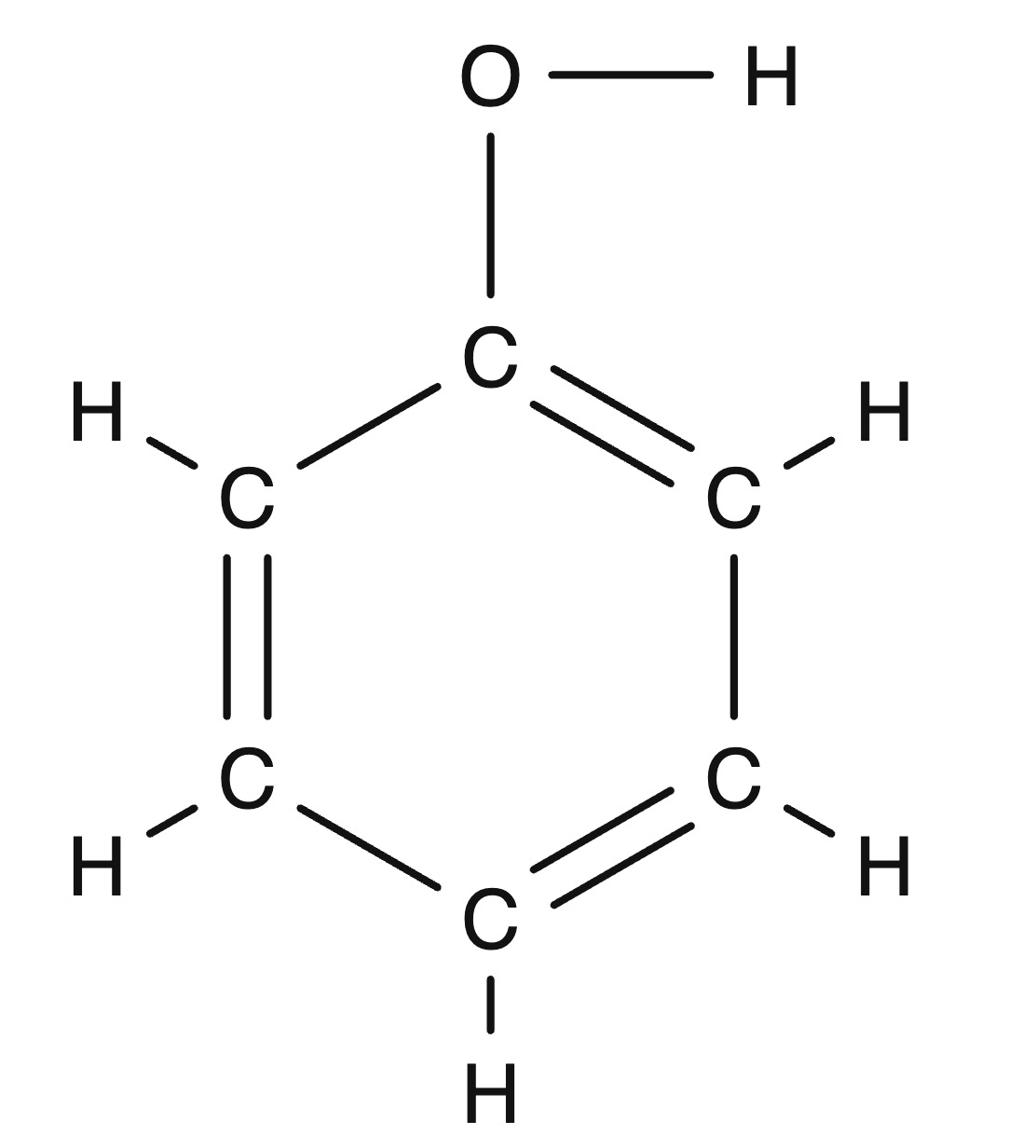}
    \includegraphics[width=\gapwidth\linewidth]{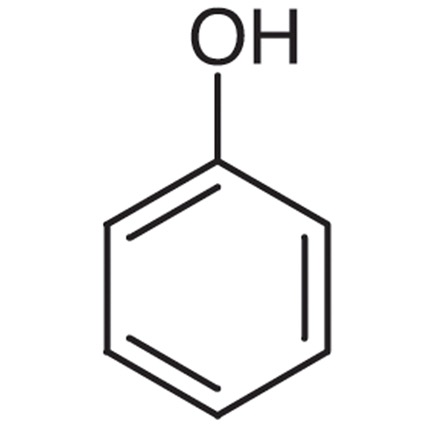}
    \caption{Two depictions of a chemical structure.}
  \end{subfigure}
  \hfill
  \begin{subfigure}{0.32\linewidth}
    \includegraphics[width=\gapwidth\linewidth]{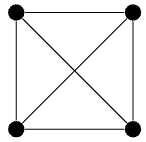}
    \includegraphics[width=\gapwidth\linewidth]{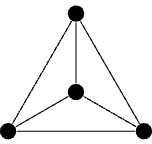}
    \caption{A graph drawn in two different layouts.}
  \end{subfigure}
  \hfill
  
  \caption{Examples of different visual representations of identical concepts across different scientific and technical domains. Each pair of images illustrates the challenge of recognizing identical relationships or structures when presented in varied visual formats.}
  \label{fig:datasamples}
\end{figure*}

\textbf{Defining conceptualization:} Undoubtedly, the ability to generalize beyond superficial patterns lies at the heart of human-like reasoning. While out-of-distribution generalization encompasses various challenges, we focus specifically on models' ability to understand different representations of the same underlying concept. To precisely characterize this aspect, we use the term "Conceptualization".

Conceptualization is a specific subset of out-of-distribution (OOD) generalization that examines a model’s ability to recognize and reason about OOD data whose \textit{underlying concept} aligns with the training distribution, but whose \textit{representation} diverges in form, structure, or manifestation.

Unlike broader OOD generalization challenges—which may involve shifts in spurious correlations, contextual biases, or surface-level features—conceptualization isolates the capacity to abstract and transfer core conceptual invariants across representational variations. (Figure~\ref{fig:datasamples})

For instance, while understanding different accents in speech recognition represents an OOD challenge, it falls outside the scope of conceptualization, as the core linguistic constructs remain consistent. Indeed, we would not attribute an American's difficulty in understanding a Scottish accent to a lack of conceptual understanding of English.

\textbf{Why Graphs?} Among the various visual representations, graphs, consisting of nodes and edges, offer an ideal test case for studying conceptualization in AI systems. While these graphical representations are easily interpreted by humans, particularly when the graphs are simple and have a limited number of nodes, they present a challenge for AI models - the ability to abstract the underlying structure regardless of visual layout.

It is worth noting that while computational algorithms exist for solving graph problems through matrix operations or adjacency list traversal, our focus here is fundamentally different. The challenge lies not in developing graph-theoretic solvers, we aim to evaluate whether AI models can solve these problems using only visual reasoning and conceptual abstraction, without relying on algorithmic computation. This distinction is crucial to align more closely with human-like reasoning.

This insight led to the creation of the ``Visual Graph Arena", a dataset and benchmark consisting of six graph-based tasks. These tasks are designed to evaluate and enhance the visual reasoning and conceptualization capabilities of AI models through various challenges. The datasets contain a diverse range of training samples, from 27,000 to 150,000, ensuring robust training environments. The graphs have 8-9 nodes, small enough such that is easy to visually inspect their properties, but still large enough to create enough samples for the datasets.

The tasks in the Visual Graph Arena revolve around three main concepts: isomorphism, path, and cycle, with two subtasks for each concept. The training sets are drawn using specific layouts (Kamada-Kawai or random, depending on the task), while the test sets employ different layouts to assess whether models can learn and apply the concepts, regardless of their visual presentation, similarly to the human capabilities.

\begin{figure*}[t]
  \centering
  
  \begin{subfigure}{0.32\linewidth}
    \includegraphics[width=\linewidth]{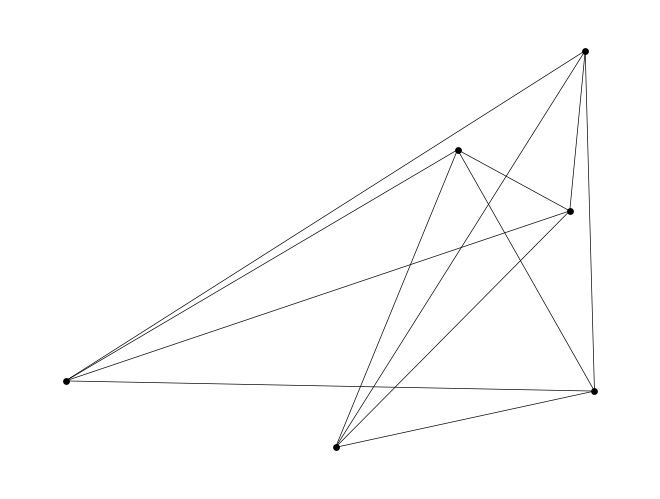}
    \caption{Random Layout}
  \end{subfigure}
  \hfill
  \begin{subfigure}{0.33\linewidth}
  \includegraphics[width=\linewidth]{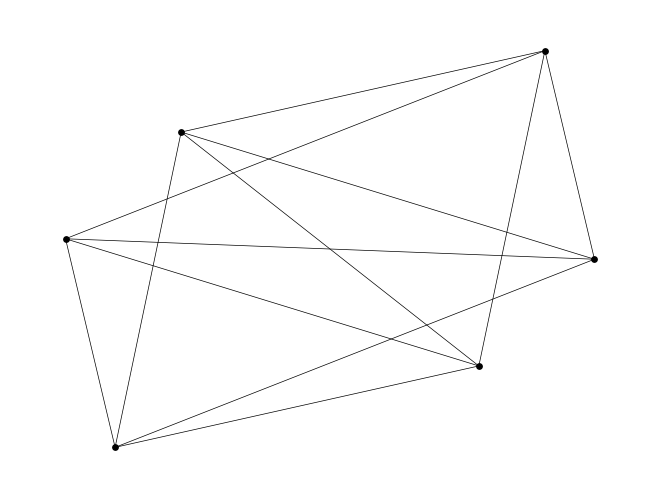}
    \caption{Kamada-Kawai Layout}
  \end{subfigure}
  \hfill
  \begin{subfigure}{0.33\linewidth}
    \includegraphics[width=\linewidth]{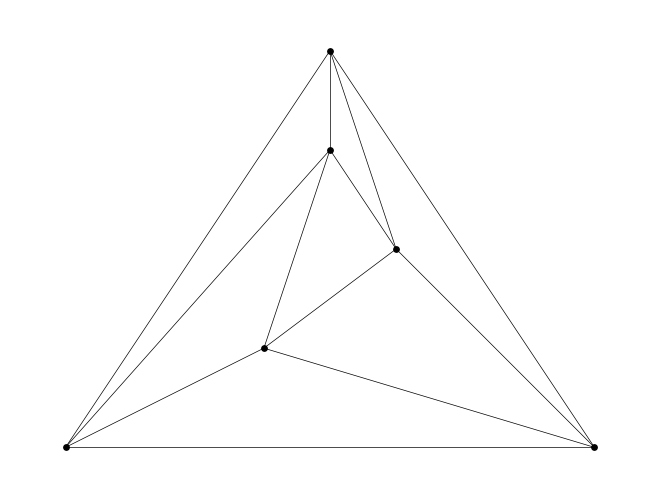}
    \caption{Planar Layout}
  \end{subfigure}
  \hfill
  
  \caption{Illustration of a single graph displayed in three distinct visualization layouts to highlight the impact of graphical representation on perceived structure and complexity.}
  \label{fig:datasamples2}
\end{figure*}

We evaluated several leading vision models,  including ViT~\cite{imageWorth16Words}, Swin Transformers~\cite{swinT}, and ConvNeXt~\cite{convnext}, on the Visual Graph Arena datasets. The results were very revealing: All models failed on the graph isomorphism detection tasks. Additionally, these models underperformed relative to the human capabilities on the other tasks, when tested with graphs drawn differently than the layouts in their training sets. Our investigation also revealed a fundamental reasoning gap in current state-of-the-art vision and multimodal language models. When we tested advanced AI models, such as GPT-o1 of OpenAI and Claude-3.5-Sonnet of Anthropic, on tasks that humans handle intuitively based on visual information alone, the results were striking. In a graph isomorphism detection task, where pairs of graphs were drawn in different layouts (Kamada-Kawai~\cite{kamadakawai} and planar), all tested models failed to recognize the identical underlying structures, while human subjects achieved more than 90\% accuracy. This stark contrast between human and AI performance highlights a critical challenge in AI conceptualization.

The Visual Graph Arena dataset helps identify AI systems' limitations in processing conceptual visual data while providing a foundation for improvement. By testing models' conceptual reasoning abilities, we work to bridge the gap between human and machine interpretation of abstract graphics.

\section{Related Work}

\textbf{Graph Datasets.} Graph datasets are a cornerstone in numerous research domains~\cite{hu2020open}, ranging from social network analysis to protein micro molecules~\cite{morris2020tudataset}. These datasets are typically divided into two main types: those that represent real-world information in graph form, like social media connections or road networks, and synthetic graph datasets crafted for specific computational experiments.

Among synthetic datasets, the variety is broad. For instance, various works have introduced datasets aimed at tasks like counting substructures within graphs~\cite{chen2020can} or distinguishing between different families of non-isomorphic graphs~\cite{wang2023towards, abboud2020surprising, balcilar2021breaking, murphy2019relational}. These synthetic datasets are primarily designed to benchmark and improve graph neural networks (GNNs). They often include data of nodes, edges, and additional features to facilitate computational processing and analysis tailored to GNN architectures.

However, a gap exists in these offerings concerning visual interpretation. Most existing graph datasets are structured for algorithmic or GNN consumption and do not address the challenge of visual graph reasoning directly. They are optimized for models that process graphs as textual or abstract data structures rather than as visual entities.

\textbf{Visual Reasoning Datasets.}  Recent advancements in machine learning have seen a burgeoning interest in tasks that interrogate visual scenes through question-answering (QA) formats~\cite{johnson2017clevr, antol2015vqa}. 

Visual QA tasks have driven progress in multimodal reasoning, with datasets like CLEVR (spatial/relational reasoning) and FigureQA (scientific plots) using synthetic data to isolate challenges~\cite{johnson2017clevr, kahou2017figureqa}.



Recent benchmarks address diverse aspects: long-context document understanding (MMLONGBENCH-DOC), chain-of-thought reasoning (Visual CoT), compositional robotic reasoning (ClevrSkills), mathematical Olympiad problems (Children’s Math Olympiads evaluation), commonsense riddles (Visual Riddles), and multimodal math reasoning (MATH-Vision)~\cite{gong2024uknow, shao2024visual, guetta2024visual, wang2024measuring, ma2024mmlongbenchdoc, cherian2024evaluating, haresh2024clevrskills}. A recent study introduces Vision Graph, a benchmark for visual reasoning in graph theory problems ~\cite{li2024visiongraph}, with two key limitations: its circular graph layouts become visually ambiguous with increasing nodes, and its dataset size is insufficient for training data-intensive models.




Our work introduces the "Visual Graph Arena" , a graph-based dataset designed not only to evaluate AI model capabilities in solving graph theory problems but also, at a higher level, to assess their conceptualization ability. This is achieved by systematically varying graph layouts during both training and testing, thereby avoiding overfitting to specific patterns and isolating reasoning challenges from perceptual artifacts. By emphasizing scalability (27,500–155,718 samples per task) and layout diversity, we ensure that all samples remain visually solvable while probing deeper cognitive processes. This focus on practical solvability—validated through human evaluation—distinguishes our benchmark as a robust tool for advancing visual graph understanding and fostering models capable of generalizing across different visual contexts.



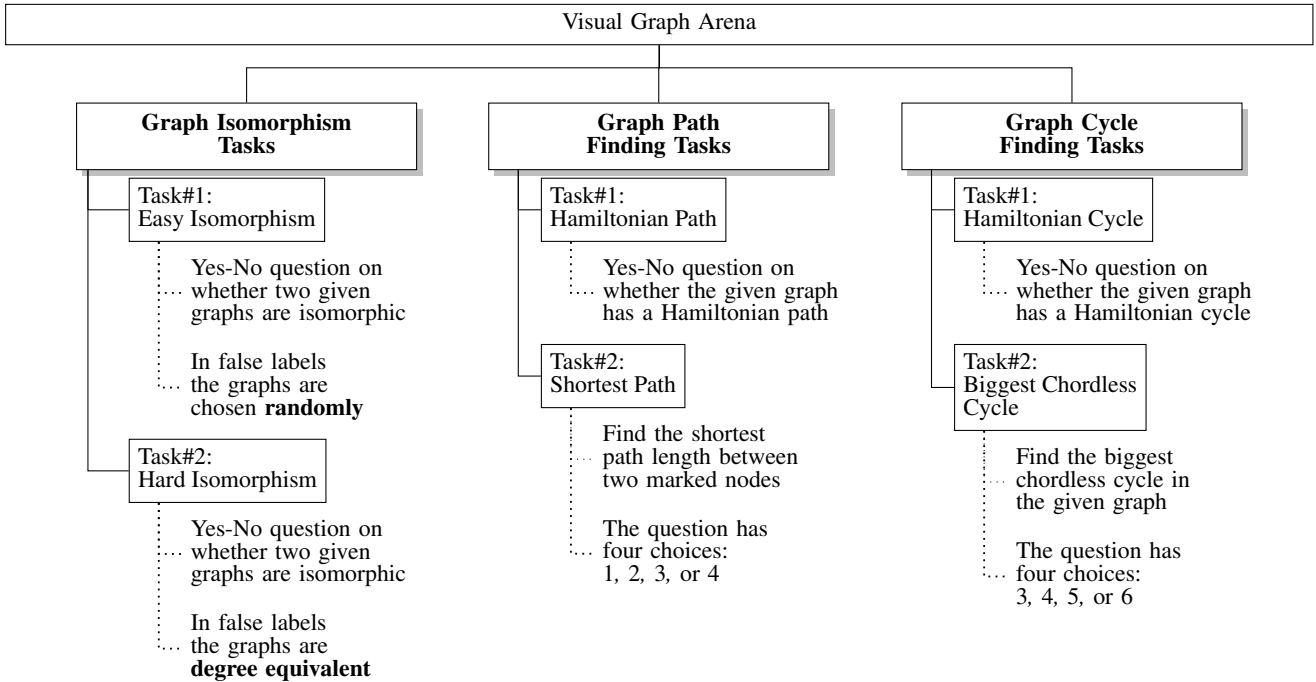
\begin{figure*}[ht]
\centering
\begin{forest}
    for tree={
        draw,
        font=\small\linespread{0.84}\selectfont,
        calign=edge midpoint,
        align=center,
if level =1{draw,
            minimum height=5.6ex,
            edge path={\noexpand\path[\forestoption{edge}]
            (!u.south) -- ++ (0,-3mm) -| (.child anchor);}
            }{},
if level>=1{grow'=0,
            folder,
            folder indent=4mm,
            l sep=7mm,
            s sep=1mm}{},
if level =2{draw, align=left}{},
if level>=3{draw=none,
           text width=11em,
           align=left}{},
              },
[Visual Graph Arena, text width=\textwidth, text centered
    [\textbf{Graph Isomorphism} \\ \textbf{Tasks}, fill=white, drop shadow, text width=0.25\textwidth, text centered
        [Task\#1: \\ Easy Isomorphism
            [Yes-No question on \\ whether two given \\ graphs are isomorphic, edge={dotted, line width=0.7pt}]
            [In false labels \\ the graphs are \\ chosen \textbf{randomly}, edge={dotted, line width=0.7pt}]
        ]
        [Task\#2: \\ Hard Isomorphism
            [Yes-No question on \\ whether two given \\ graphs are isomorphic, edge={dotted, line width=0.7pt}]
            [In false labels \\ the graphs are \\ \textbf{degree equivalent}, edge={dotted, line width=0.7pt}]
        ]
    ]
    [\textbf{Graph Path} \\ \textbf{Finding Tasks}, fill=white, drop shadow, text width=0.25\textwidth, text centered
        [Task\#1: \\ Hamiltonian Path
           [Yes-No question on \\ whether the given graph\\has a Hamiltonian path, edge={dotted, line width=0.7pt}]
        ]
        [Task\#2: \\ Shortest Path
            [Find the shortest\\path length between\\two marked nodes, edge=dotted]
            [The question has\\four choices:\\1\textit{,} 2\textit{,} 3\textit{,} or 4, edge={dotted, line width=0.7pt}]
        ]
    ]
    [\textbf{Graph Cycle} \\ \textbf{Finding Tasks}, fill=white, drop shadow, text width=0.25\textwidth, text centered
        [Task\#1: \\ Hamiltonian Cycle
           [Yes-No question on \\ whether the given graph\\has a Hamiltonian cycle, edge={dotted, line width=0.7pt}]
        ]
        [Task\#2: \\ Biggest Chordless\\Cycle
            [Find the biggest\\chordless cycle in\\the given graph, edge=dotted]
            [The question has\\four choices:\\3\textit{,} 4\textit{,} 5\textit{,} or 6, edge={dotted, line width=0.7pt}]
        ]
    ]
]
\label{dataset_overview}
\end{forest}
\caption{Overview of the 'Visual Graph Arena Benchmark' detailing the division of tasks into categories of Graph Isomorphism, Graph Path Finding, and Graph Cycle Finding. Each category further breaks down into two subtasks.}
\label{fig:tree}
\end{figure*}


\section{Visual Graph Arena}

This section introduces the Visual Graph Arena Dataset, meticulously crafted to challenge and evaluate the visual reasoning capabilities of AI models through a series of graph-based tasks. The dataset is structured around three primary concepts, each divided into two tasks, resulting in a total of six distinct datasets. These datasets are designed to rigorously test various aspects of graph understanding and reasoning.

In order to ensure that the graphs are both manageable and visually interpretable, we have constrained the number of nodes to range between 8 and 9. This node count strikes a balance between maintaining visual simplicity and providing sufficient complexity for generating robust training and test sets, which are suitable for deep learning applications.

The tasks are devised around core graph concepts that are intuitively recognizable by humans when interacting with small graphs:

\begin{enumerate}
    \item \textbf{Graph Isomorphism:} Tests the ability to determine the isomorphic nature of graph pairs.
    \item \textbf{Path Finding:} Focuses on the capability to identify various path types within the graphs.
    \item \textbf{Cycle Finding:} Challenges the models to detect different types of cycles.
\end{enumerate}

Figure~\ref{fig:tree} shows an overview of the Visual-Graph-Arena Benchmark tasks and datasets, and Table~\ref{tab:sizes} shows the breakdown of the tasks with the number of samples in the training and test set for each task.

In the subsequent subsections, we will describe each task in detail.

\subsection{Graph-Isomorphism Tasks}

This task focuses on graph isomorphism, utilizing graphs consisting of 8-9 nodes, to test the models' capabilities in recognizing isomorphic structures. Graph isomorphism refers to the concept where two distinct graph representations share the same number of vertices, edges, and connectivity patterns, essentially making them indistinguishable in graph-theoretic terms despite their different visual layouts. To assess this, the task presents pairs of graphs drawn in different layouts, one in Kamada-Kawai and the other in planar, and models must determine whether the graphs are isomorphic.

The task is structured to ensure a balanced approach: each set contains an equal number of isomorphic and non-isomorphic graph pairs. This balance is crucial to prevent models from learning and exploiting label distribution biases instead of mastering the intended isomorphic detection.

\textbf{Difficulty Levels:} The graph-isomorphism-recognition task demands quite sophisticated reasoning from the models, as superficial attributes like edge count or node degree are insufficient for determining isomorphism. The models must engage in deeper analytical processes to also discern underlying structural equivalences or differences.

\textbf{Easy:} In the easier version of this task, the second graph in each pair is selected randomly from the set of all possible graphs with 8-9 nodes. Given the vast diversity within this family, non-isomorphic pairs often exhibit clear, easily discernible differences, such as varying node degrees, simplifying the detection of non-isomorphism.

\begin{figure}[H]
    \centering
    \vspace*{1mm}
    \begin{tikzpicture}[scale=1.1]
    \foreach \i in {1,...,6} {
    \node[circle,fill,inner sep=2pt] (n\i) at (360/6*\i:1) {};
    }
    \draw (n1) -- (n2) -- (n3) -- (n4) -- (n5);
    \draw (n6) -- (n1) -- (n3);
    \end{tikzpicture}
    \quad
    \begin{tikzpicture}[scale=1.1]
    \foreach \i in {1,...,6} {
    \node[circle,fill,inner sep=2pt] (n\i) at (360/6*\i:1) {};
    }
    \draw (n1) -- (n2) -- (n3) -- (n4) -- (n5);
    \draw (n6) -- (n1) -- (n4);
    \end{tikzpicture}
    \caption{Two non-isomorphic \emph{degree-equivalent} graphs with the degree sequence of \{3, 3, 2, 2, 1, 1\}.}
    \label{fig:6node_graphs}
\end{figure}
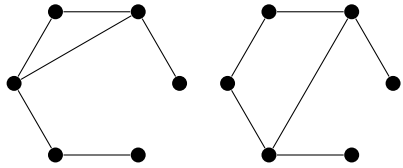

\textbf{Hard:} The hard category intensifies the challenge by selecting graph pairs that are degree-equivalent. These pairs are visually different but share identical numbers of edges and identical node degrees. Figure~\ref{fig:6node_graphs} shows an example of a non-isomorphic degree-equivalent pair of graphs. 

These varied levels of difficulty are designed to progressively train and evaluate the AI's depth of understanding and its ability to perform complex visual reasoning, mirroring the cognitive processes humans use in interpreting graphical data.

\subsection{Graph-Path-Finding Tasks}
The Graph-Path-Finding datasets explore the model's ability to understand the concept of a path within a graph, where each graph contains 8-9 nodes. The task is divided into two subtasks:

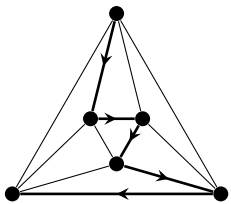
\begin{figure}[h]
    \centering
    \begin{tikzpicture}[scale=0.8]
            \tikzset{mid arrow/.style={
                decoration={markings, mark=at position 0.5 with {\arrow{stealth}}},
                postaction={decorate}
            }}
            
        \foreach \i in {1,...,3} {
        \node[circle,fill,inner sep=2pt] (n\i) at (90 + 360/3*\i:2) {};
        }
        \foreach \i in {1,...,3} {
        \node[circle,fill,inner sep=2pt] (m\i) at (30 + 360/3*\i:0.5) {};
        }
        \draw (m2) -- (n1) -- (n2) -- (n3) -- (n1) -- (m1);
        \draw (n3) -- (m3) -- (m2) -- (m1) -- (m3) -- (n2);
        \draw[line width=1pt, mid arrow] (n3) -- (m1);
        \draw[line width=1pt, mid arrow] (m1) -- (m3);
        \draw[line width=1pt, mid arrow] (m3) -- (m2);
        \draw[line width=1pt, mid arrow] (m2) -- (n2);
        \draw[line width=1pt, mid arrow] (n2) -- (n1);
    \end{tikzpicture}
    \caption{Example of a Hamiltonian path, a path that visits each node a single time.}
    \label{fig:hampath}
\end{figure}

\textbf{Haminltonian Path:} The objective is to determine whether a graph contains a Hamiltonian path—a path that visits each vertex exactly once. Figure~\ref{fig:hampath} shows an example of such a path in a graph. This is essentially a binary classification problem, with labels indicating the presence or absence of a Hamiltonian path. To maintain balance, the dataset includes an equal number of graphs with and without such paths. The training-set graphs are drawn in Kamada-Kawai layout, and the test-set graphs are drawn in planar layout. This variation in the training and testing layouts helps to assess the conceptualization ability of the models.

\textbf{Shortest Path:} This task presents graphs in which two nodes are specifically marked. The goal is to find the length of the shortest path between these nodes, with potential answers of 1, 2, 3, or 4. This setup frames the task as a classification problem with four classes. The training dataset contains graphs drawn from the random layout. The test set samples are drawn from random, Kamada-Kawai, and planar layouts, forming three testing variations. 



\subsection{Graph-Cycle-Finding Tasks}
These tasks and associated datasets test the ability of a model to understand the notion of cycles within graphs, where each graph consists of 8-9 nodes. As before, these tasks also feature two levels of difficulty, to assess different aspects of the models' capabilities.

\textbf{Hamiltonian Cycle:} The goal here is to determine whether the graph contains a Hamiltonian cycle, that is, a cycle which visits each vertex exactly once. The dataset is balanced between graphs that do and do-not contain a Hamiltonian cycle. To evaluate the models' ability to generalize their learning, similar to the Hamiltonian path task, the training set graphs are visualized in the Kamada-Kawai layout, whereas the test set is drawn in a planar layout.

\begin{figure}[h]
    \centering
    \begin{tikzpicture}[scale=1.1]
        \foreach \i in {1,...,6} {
        \node[circle,fill,inner sep=2pt] (n\i) at (360/6*\i:1) {};
        }
        \node[circle,fill,inner sep=2pt] (mid) at (0:0) {};
        \node[circle,fill,inner sep=2pt] (m2) at (180:2) {};
        \node[circle,fill,inner sep=2pt] (m1) at (150:1.7) {};
        \node[circle,fill,inner sep=2pt] (m3) at (-150:1.7) {};
        \draw (n4) -- (m3) -- (m2) -- (m1) -- (n2);
        \draw (m1) -- (m3);
        \draw (n5) -- (mid);
        \draw (n2) -- (mid) -- (n4);
        \draw (n1) -- (n2) -- (n3) -- (n4) -- (n5) -- (n6) -- (n1);
        \draw[line width=1pt, red]  (n1) -- (n2) -- (m1) -- (m3) -- (n4) -- (n5) -- (n6) -- (n1);
        \end{tikzpicture}
            \caption{An example for the biggest-chordless-cycle problem in a graph with 10 nodes.}
            \label{fig:chordless}
\end{figure}
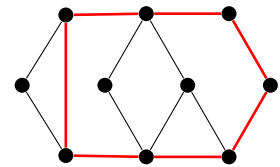

\begin{table*}[]
  \caption{Number of samples in the training and test sets of the Visual Graph Arena tasks.}
  \label{tab:samples}
  \vspace*{1mm}
  \setlength{\tabcolsep}{5pt} 
  \centering
  \begin{tabular}{@{}lcccccc@{}}
     &\multicolumn{2}{c}{\textbf{Isomorphism}} & \multicolumn{2}{c}{\textbf{Path}} & \multicolumn{2}{c}{\textbf{Cycle}} \\
     \cmidrule(r){2-3} \cmidrule(lr){4-5} \cmidrule(l){6-7}
     &Easy&Hard& Hamiltonian&Shortest&Hamiltonian&Biggest Chordless\\
     \midrule
    \emph{\# Training Samples} & 140,000 & 127,374 & 25,000 & 80,000 & 69,935 & 80,000 \\
    \emph{\# Testing Samples} & 15,718 & 14,298 & 2,480 & 8,672 & 7,740 & 6,484 \\
  \end{tabular}
  \label{tab:sizes}
\end{table*}

\textbf{Biggest Chordless Cycle:} In this challenging scenario, the task involves identifying the size of the largest chordless cycle in a graph. A chordless cycle is defined as one where no two vertices of the cycle are connected by an edge that is not part of the cycle itself. Figure~\ref{fig:chordless} shows an example of the biggest chordless cycle in a graph with 10 nodes. The possible answers are 3, 4, 5, or 6, effectively making this a classification task with four options. Similar to the previous task, the training and testing samples are drawn in the Kamada-Kawai and planar layouts, respectively.


The detailed process of the creation of the Visual-Graph-Arena datasets and example samples of each of the tasks they include is presented in the Appendix.

\begin{table*}[ht]
\centering
\caption{Comparative performance of humans, vision models, and MLLMs on Visual Graph Arena tasks. Human participants achieved near-perfect accuracy (88.2-100\%), excelling in shortest path identification. Vision models failed isomorphism tasks with ConvNeXt outperforming transformer-based models in non-failing tasks. MLLMs showed near-random performance except GPT-o1, which achieved limited success in shortest path (55\%) and Hamiltonian cycle (66.6\%) tasks.}
\vspace{10pt}
\begin{threeparttable}
\begin{tabular}{@{}lcccccccc@{}}
\setlength{\tabcolsep}{0pt} 
\centering
& \multicolumn{2}{c}{\textbf{Isomorphism}} & \multicolumn{4}{c}{\textbf{Path}} & \multicolumn{2}{c}{\textbf{Cycle}} \\
\cmidrule(r){2-3} \cmidrule(lr){4-7} \cmidrule(l){8-9}
& easy & hard & hamiltonian\tnote{1} & \multicolumn{3}{c}{shortest path\tnote{2}} & hamiltonian\tnote{1} & chordless\tnote{1} \\ \cmidrule{5-7}
& & & & random & kawai & planar & & \\
\midrule
\rowcolor{metablue!10} human & 95.0 & 91.6 & 96.6 & 98.3 & 100 & 100 & 93.3 & 88.2 \\ \addlinespace[0.6em]
Vit Base & FAIL\tnote{3} & FAIL & 57 & 67.4 & 66.7 & 64.4 & 67.8 & 32.3 \\ \addlinespace[0.3em]
Swin-T Base & FAIL & FAIL & 65.8 & 68.6 & 65.7 & 65.4 & 71.1 & 34.4 \\ \addlinespace[0.3em]
ConvNext Base & FAIL & FAIL & 72.9 & 73.3 & 82.4 & 73.3 & 74.9 & 36.3 \\ \addlinespace[0.6em]
SigLIP Base	& \cellcolor{metablue!10}54.4	&FAIL	&59.5	& 25.2 & - &-	&63.5	&28.0 \\ \addlinespace[0.2em]
DINov2	Base &FAIL	&FAIL	&56.8	& 36.4 & -&-	&FAIL	&31.1 \\ \addlinespace[0.6em]
GPT-o1 & FAIL & FAIL & FAIL & \cellcolor{metablue!10}55.0 & - & - & \cellcolor{metablue!10}66.0 & FAIL \\ \addlinespace[0.2em]
GPT-4o & FAIL & FAIL & FAIL & FAIL & - & - & FAIL & FAIL \\ \addlinespace[0.2em]
Claude 3.5 Sonnet & FAIL & FAIL & FAIL & FAIL & - & - & FAIL & FAIL
\end{tabular}

\begin{tablenotes}
\item[1] These tasks are trained on \emph{kawai} layout and tested on \emph{planar} layout.
\item[2] Shortest path task is trained on \emph{random} layout, and tested on \emph{random}, \emph{kawai} and \emph{planar} layouts.
\item[3] We use FAIL to indicate near-random performance, borrowing the term from~\cite{tay2021long}.
\end{tablenotes}
\end{threeparttable}
\label{table:performance}
\end{table*}

\section{Experiments}
In order to rigorously evaluate the effectiveness of Visual Graph Arena, we conducted a series of experiments, aimed at benchmarking both human and AI performance on the graph-based tasks. Initially, we established a baseline by testing human subjects, providing a comparative standard for assessing AI capabilities in similar conditions. Following this, we moved on to evaluate advanced multimodal-llms including GPT-o1 and Claude 3.5 Sonnet, to test their proficiency in visual reasoning tasks that demand an integration of visual and textual analysis. Subsequently, we focused on the performance of leading vision models, which were specifically trained on our datasets and then tested on the designated test sets containing a different graph layout.

We detail each experiment, and discuss the methodologies employed, the models tested, and the results obtained. 

\textbf{Humans:} To establish a baseline of human performance on the VGA tasks, we conducted experiments with 15 subjects with a total of 24 questions, distributing four questions per task. Participants were drawn from the student and staff population of the engineering faculty, possessing a general understanding of graph concepts, though not specifically in graph isomorphism, Hamiltonian paths and cycles, or chordless cycles—most requested definition and clarity of these concepts. More detail is provided in the Appendix. 

\textbf{Vision Models:} We tested three categories of vision models:
\begin{itemize}
\vspace{-7pt}
\setlength{\itemsep}{0pt}
  \item \textit{Supervised Pre-trained Models}: ViT, ConvNeXt, and Swin-T Base, each pre-trained on the ImageNet dataset using supervised learning with labeled data.
  \item \textit{Multimodal Vision-Language Model}: SigLIP Base, pre-trained on image-text pairs.
  \item \textit{Self-Supervised Vision Model}: DINOv2 Base, trained on unlabeled image data.
\vspace{-7pt}
\end{itemize}

All models were fine-tuned on our tasks across 10 epochs. Details regarding the training parameters, settings, and additional considerations are provided in the Appendix.

\textbf{MLLMs:} We evaluated GPT-4o and o1, Claude 3 Opus and 3.5 Sonnet, and Google Gemini on 100 random samples per task to determine their ability to handle graph-based visual reasoning. Unfortunately, all models performed poorly across tasks, with o1 being the only exception, showing higher than random performance in shortest path and Hamiltonian cycle problems.

\subsection{Results}


\begin{table}[h]
  \caption{Confusion matrix for GPT-o1 performance on the shortest path task. Rows represent ground-truth shortest path lengths (1–4), columns show model predictions. Values indicate prediction percentages (diagonal = correct classifications). The model performs best on length 2 (69\% accuracy) but frequently confuses lengths 1→2 (67\%), suggesting difficulty distinguishing adjacent node path lengths.)}
  \label{tab:confusion}
  \vspace*{1mm}
  \setlength{\tabcolsep}{8pt}
  \centering
  \begin{tabular}{@{}ccccc@{}}
    \diagbox{Label}{Output} & 1 & 2 & 3 & 4 \\
    \midrule
    Length 1 & \cellcolor{metablue!28}28.0 & \cellcolor{metablue!67}67.0 & \cellcolor{metablue!5}5.0 & \cellcolor{metablue!0}0.0 \\
    Length 2 & \cellcolor{metablue!18}18.0 & \cellcolor{metablue!69}69.0 & \cellcolor{metablue!13}13.0 & \cellcolor{metablue!0}0.0 \\
    Length 3 & \cellcolor{metablue!0}0.0 & \cellcolor{metablue!40}40.0 & \cellcolor{metablue!60}60.0 & \cellcolor{metablue!0}0.0 \\
    Length 4 & \cellcolor{metablue!0}0.0 & \cellcolor{metablue!5}5.0 & \cellcolor{metablue!40}40.0 & \cellcolor{metablue!55}55.0 \\
    \bottomrule
  \end{tabular}
  \label{tab:confusion_matrix}
\end{table}

Table~\ref{table:performance} presents the results of the experiments conducted on human subjects, vision models, and MLLMs for the Visaul Graph Arena tasks. The human participants demonstrated strong performance across all tasks, achieving 100\% accuracy in finding the shortest path length and over 90\% accuracy in the other tasks. Finding the length of the biggest chordless cycle proved to be the most challenging for human subjects, with an accuracy of 88.2\%, which was their lowest performance among all tasks.

For the vision models, we report the best validation accuracy achieved during the training epochs. Notably, all models except SigLIP failed in the isomorphism tasks, with SigLIP achieving 54.4\% accuracy on the easy isomorphism task while still failing on the hard variant. This indicates a significant limitation in most models' ability to reason about graph isomorphism from visual inputs. In the other tasks, the models exhibited relatively poor performance, considering that the labels were either binary or limited to four choices.
The chordless cycle task emerged as the most challenging for all vision models, with the ConvNeXt model achieving the highest accuracy of 36.3\%. An interesting observation is that the ConvNeXt model consistently outperformed the Vision Transformer (ViT) and Swin Transformer (Swin-T) models on all tasks where the models did not completely fail. This performance gap was most significant in the task of finding the shortest path length in the Kawai layout, where ConvNeXt achieved an accuracy of 82.4\%, surpassing ViT and Swin-T by 17.7 and 16.7 percentage points, respectively. The newer models, SigLIP and DINov2, showed mixed results, with SigLIP demonstrating some capability in isomorphism detection but both models generally underperforming compared to ConvNeXt. This finding suggests that convolutional architectures may be more effective than transformer-based models in capturing and reasoning about visual concepts in graph representations.

\begin{figure}[t]
    \centering
    \includegraphics[width=0.47\textwidth]{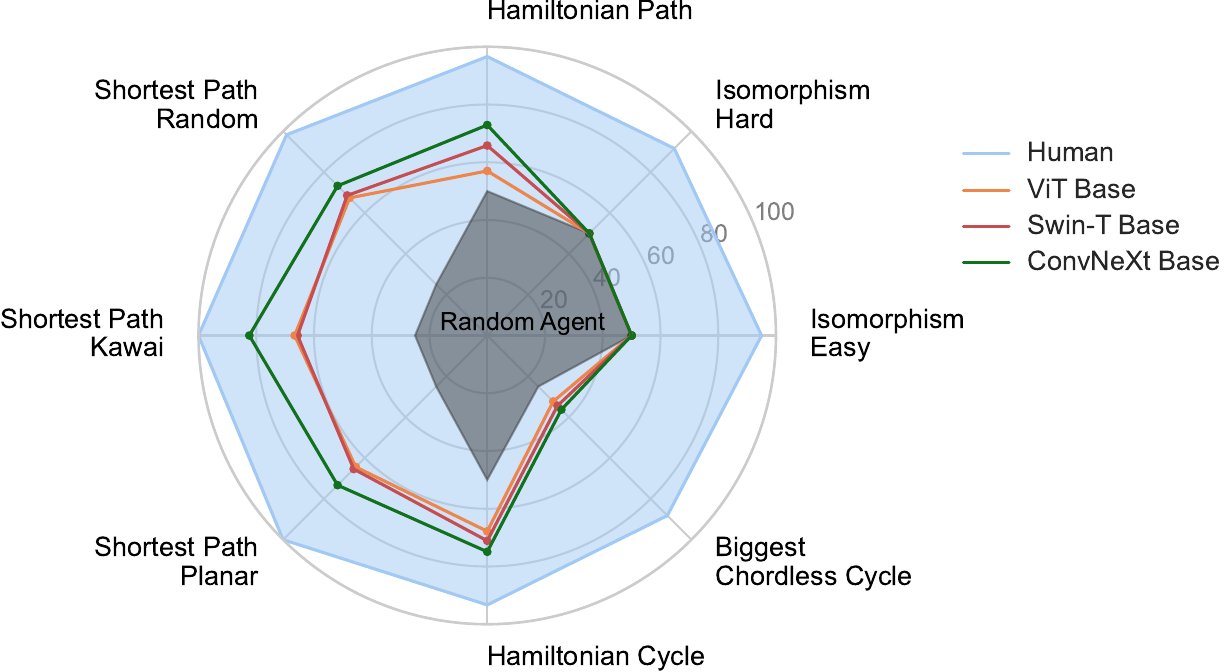}
    \caption{Breakdown of the performance comparison of the VIT, Swin, and ConvNeXt models to humans on the Visual Graph Arena tasks. The central area displays the performance of a random agent.}
    \label{fig:radar}

\end{figure}

Figure~\ref{fig:radar} illustrates the performance comparison of the trained models on the tasks of the VGA benchmark. The performance of a random agent is in gray. On the graph isomorphism tasks all models performed like a random agent.

Furthermore, we evaluated the performance of advanced MLLMs on the Visual Graph Arena tasks. Quite surprisingly, most tested MLLMs, including GPT-4o, Claude 3.5 Sonnet, 3 Opus, and Google Gemini, failed to solve any of the given tasks. However, GPT-o1 demonstrated partial success, achieving non-random accuracy on two specific tasks: shortest path (55\% accuracy) and Hamiltonian cycle (67\% accuracy). Our analysis reveals that GPT-o1's above-random performance on the Hamiltonian cycle task is primarily attributed to the model's ability to identify graphs with leaf nodes, which trivially cannot contain Hamiltonian cycles. When cases involving leaf nodes are excluded, GPT-o1's performance drops to near-random levels, which may also explain its failure on the related Hamiltonian path task. This suggests that the model relies on specific structural cues rather than a deeper understanding of graph theory.

For the shortest path task, GPT-o1 exhibited varying performance across different path lengths, as shown in Table~\ref{tab:confusion_matrix}. The confusion matrix highlights that the model performs best when the shortest path length is 2 (69\% accuracy) but frequently misclassifies adjacent nodes, such as confusing lengths 1 and 2. These observations underscore the limitations of current MLLMs in handling basic graph-reasoning tasks, particularly when visual reasoning on graph-structured data is required.

The results of our experiments reveal significant challenges and limitations in the current state of the art vision models and MLLMs when it comes to reasoning about graph-structured visual data. While human participants exhibit strong performance on the Visual Graph Arena benchmark tasks, the AI models struggle to match human-level reasoning capabilities. These findings emphasize the need for further research and development of AI systems that can effectively understand and reason about abstract graphical representations, bridging the gap between human-like visual understanding and current AI capabilities. Our Visual Graph Arena may help to overcome these limitations.

\section{Conceptualization Analysis in GPT-o1}
In our evaluation, we observe behavioral patterns in MLLMs, particularly o1, that fundamentally diverge from mature human intelligence, that questions the capability of these models to conceptualize. Specifically, we identified two distinct anomalies.

\subsection{The Middle-Score Anomaly}
The Middle-Score Anomaly emerges in atomic-task assessments of intelligent agents where only extreme performances (perfect or minimal) are expected, while intermediate performance is anomalous.

Consider identifying numbers ending in zero—an atomic task. For a human, we expect performance to be near either 100\% or near 50\% (indicating guessing, not having task knowledge). An 80\% score would be anomalous because this task cannot be partially known—one either knows to check the last digit or doesn't.

\subsection{The Easier-Worse Anomaly}
The Easier-Worse Anomaly emerges in atomic-task assessments of intelligent agents where performance on simpler instances is unexpectedly inferior to performance on more complex variants of the same task. This pattern contradicts fundamental assumptions about cognitive capabilities.

Consider finding the shortest path between two nodes in a graph. For genuine intelligence, performing considerably better at finding shortest paths with 3 or 4 length than finding shortest path between adjacent nodes would be anomalous. Such inverted performance within a single atomic task cannot be reconciled with genuine understanding.

\subsection{From Anomalies to Pseudo-Intelligence}
Our experiments with GPT-o1 strongly demonstrate both anomalies. In Table~\ref{tab:confusion_matrix}, when tasked with finding shortest paths in graphs, the model achieves 29\% accuracy for adjacent nodes (path length 1) compared to ~70\% accuracy for paths of length 2, a clear case of the Easier-Worse Anomaly.

The presence of both anomalies defies explanation in terms of low intelligence, but rather suggests the existence of a pseudo-intelligence (likely result of imitative-probabilistic reasoning)—a form of capability that mimics intelligence in its above-chance success rate while failing to demonstrate the conceptualization of genuine intelligence.

This limitation becomes particularly significant given the scale of contemporary training. As noted by Sutskever, current models have been trained on nearly the entire internet~\cite{sutskever2024neurips}, and our investigations confirm substantial exposure to graph-related content. This brings into question whether current probabilistic training approaches are inherently incapable of achieving conceptualization. Being said, other training methods like reinforcement learning, due to their foundations in biological learning processes, may serve as a path forward for solving this challenge.

\section{Conclusions}

In this paper, we introduced the Visual Graph Arena, a collection of six datasets designed designed to evaluate AI models' capacity for visual reasoning and conceptual understanding through graph-based tasks. By focusing on isomorphism detection, path finding, and cycle analysis across diverse visual layouts, our work isolates the challenge of conceptualization—the ability to recognize invariant conceptual properties despite variations in representation.

Our experiments revealed performance gap between humans and state-of-the-art AI models. Humans achieved near-perfect accuracy (88–100\%) across tasks, while vision models struggled significantly, especially when presented with graphs drawn using different layouts. MLLMs (GPT-o1, Claude 3.5 Sonnet) struggled to perform even basic graph reasoning tasks often performing no better than random chance. This finding highlights the need for further research of AI systems that can effectively conceptualize and reason about abstract graphical representations.

The Visual Graph Arena serves as a valuable resource for the AI research community, providing a foundation for advancing the field of visual reasoning on graph-structured data. By addressing the challenges posed by this dataset, researchers can work towards developing more robust and flexible AI systems that can bridge the gap between human-like conceptual understanding and reasoning and machine interpretation of abstract graphical information.


\textbf{Future Work.} Expanding visual conceptualization benchmarks to domains like chemical structures and logic circuits, where reasoning over structural invariants is critical, represents a vital next step. While developing such datasets demands domain expertise, precise annotation, and rigorous validation to ensure fidelity, these efforts will deepen our understanding of AI’s conceptualization capabilities. Such datasets will accelerate progress toward AI systems capable of human-like generalization across representations.

\section*{Impact Statement}
This paper presents work whose goal is to advance the field of 
Machine Learning. There are many potential societal consequences 
of our work, none which we feel must be specifically highlighted here.

\section*{Acknowledgments} Zahra Babaiee and Radu Grosu are supported by the
Austrian Science Fund (FWF) project MATTO-GBM I 6605. Peyman M. Kiasari is supported by
the TU Wien TrustACPS PhD School program that is supported by TTTech Auto and
B\&C Privatstiftung.

\bibliography{references}
\bibliographystyle{icml2025}

\newpage
\appendix
\onecolumn

\section{Dataset Description}

The Visual Graph Arena (VGA) Benchmark is a collection of six datasets designed to evaluate and enhance the visual reasoning capabilities of AI models on graph-based tasks. The benchmark aims to bridge the gap between human-like understanding and machine interpretation of abstract graphical information. The datasets are intended for research purposes in the field of artificial intelligence, specifically focusing on advancing visual reasoning on graph-structured data.

The benchmark datasets are hosted on the Visual Graph Arena website at \href{http://visualgrapharena.com/}{visualgrapharena.com}. The website provides access to the datasets, along with documentation and usage guidelines.

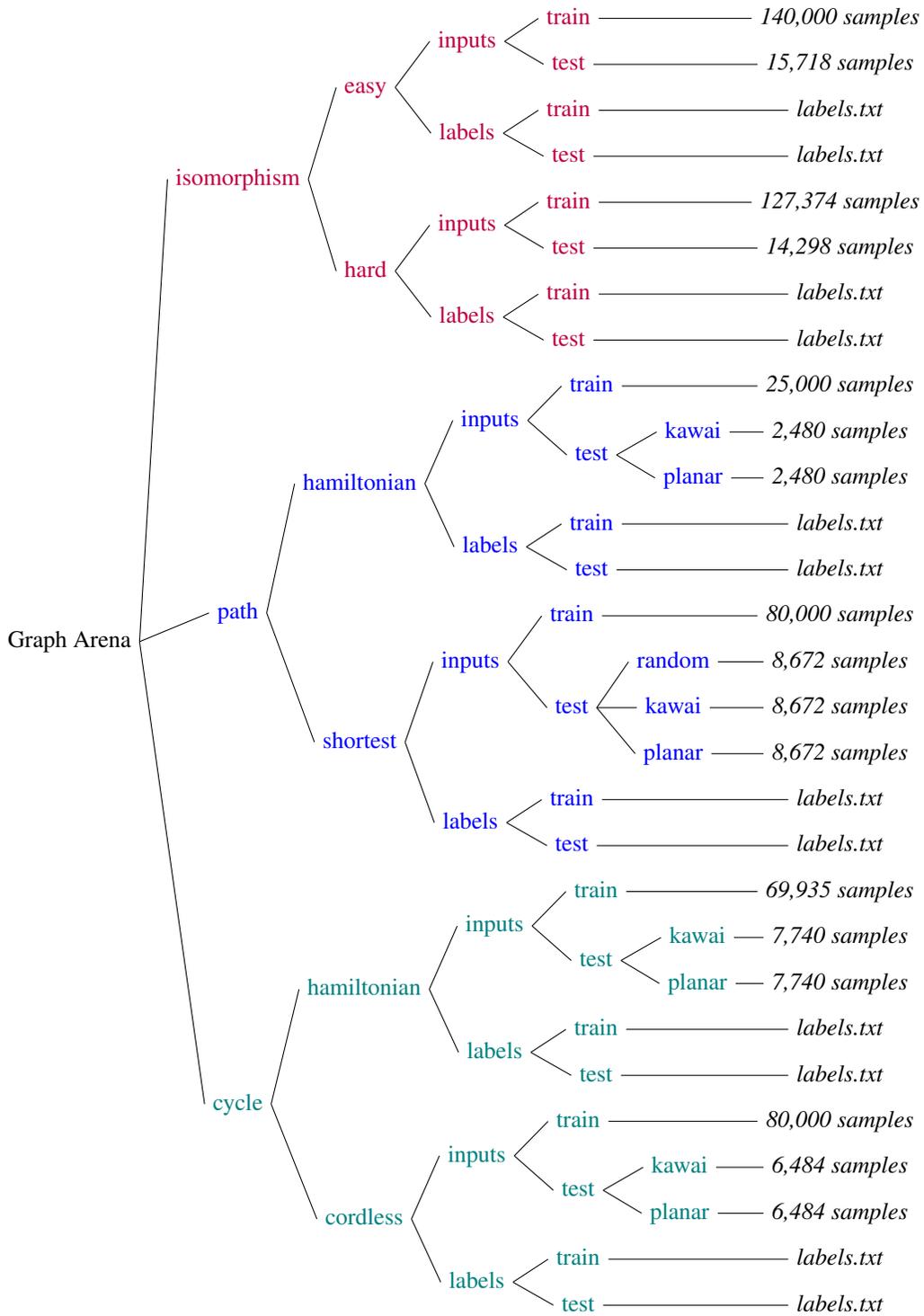
\begin{figure}[H]
\centering
\begin{forest}
    for tree={grow'=0 ,fit=band, l=1.5cm, s sep=0.01cm,
        if n children=0{
      font=\itshape,
      tier=terminal,
    }{},}
    [Graph Arena
        [isomorphism, purple
            [easy, purple
                [inputs, purple
                    [train, purple[\textit{140,000 samples}]]
                    [test, purple[\textit{15,718 samples}]]
                ]
                [labels, purple
                    [train, purple[\textit{labels.txt}]]
                    [test, purple[\textit{labels.txt}]]
                ]]
            [hard, purple
                [inputs, purple
                    [train, purple[\textit{127,374 samples}]]
                    [test, purple[\textit{14,298 samples}]]
                ]
                [labels, purple
                    [train, purple[\textit{labels.txt}]]
                    [test, purple[\textit{labels.txt}]]
                ]
            ]
        ]
        [path, blue
            [hamiltonian, blue
                [inputs, blue
                    [train, blue[\textit{25,000 samples}]]
                    [test, blue
                        [kawai, blue[\textit{2,480 samples}]]
                        [planar, blue[\textit{2,480 samples}]]
                    ]
                ]
                [labels, blue
                    [train, blue[\textit{labels.txt}]]
                    [test, blue[\textit{labels.txt}]]
                ]
            ]
            [shortest, blue
                [inputs, blue
                    [train, blue[\textit{80,000 samples}]]
                    [test, blue
                        [random, blue[\textit{8,672 samples}]]
                        [kawai, blue[\textit{8,672 samples}]]
                        [planar, blue[\textit{8,672 samples}]]
                    ]
                ]
                [labels, blue
                    [train, blue[\textit{labels.txt}]]
                    [test, blue[\textit{labels.txt}]]
                ]
            ]
        ]
        [cycle, teal
            [hamiltonian, teal
                [inputs, teal
                    [train, teal[\textit{69,935 samples}]]
                    [test, teal
                        [kawai, teal[\textit{7,740 samples}]]
                        [planar, teal[\textit{7,740 samples}]]
                    ]
                ]
                [labels, teal
                    [train, teal[\textit{labels.txt}]]
                    [test, teal[\textit{labels.txt}]]
                ]
            ]
            [cordless, teal
                [inputs, teal
                    [train, teal[\textit{80,000 samples}]]
                    [test, teal
                        [kawai, teal[\textit{6,484 samples}]]
                        [planar, teal[\textit{6,484 samples}]]
                    ]
                ]
                [labels, teal
                    [train, teal[\textit{labels.txt}]]
                    [test, teal[\textit{labels.txt}]]
                ]
            ]
        ]
    ]
\end{forest}
\caption{Overview of the directory structure of the Visual Graph Arena benchmark.}
\label{fig:graph-arena}
\end{figure}

Figure~\ref{fig:graph-arena} illustrates the directory structure of the Visual Graph Arena (VGA) dataset. The VGA dataset is structured to facilitate easy access to various graph analysis tasks including isomorphism, path problems, and cycle detection. Each category is further subdivided into training and testing datasets with detailed sample counts and label files. For each input data, 'image\_\{i\}.png' corresponds to the i-th row of it's 'label.txt'. For more information and to access the dataset, please visit the VGA dataset website at \href{http://visualgrapharena.com/}{visualgrapharena.com}. 

\textit{Author Statement:} We, the authors, take full responsibility for any potential violation of rights related to the Visual Graph Arena dataset and its materials. We confirm that the dataset is our original work and grant the hosting platform permission to distribute it under the CC BY 4.0 license. We will ensure proper access and maintenance of the dataset. 


\section{Dataset Creation}

To create the datasets, we utilized the collection of connected graphs with 8 and 9 nodes from \url{https://users.cecs.anu.edu.au/~bdm/data/graphs.html}. This collection comprises 11,117 connected graphs with 8 nodes and 261,080 connected graphs with 9 nodes. The Python library NetworkX was employed to visualize the graphs in plots for the datasets. Each sample was plotted within a square canvas of $700 \times 700$ pixels. The following details the specific steps involved in creating each task:
\subsection{Isomorphism Tasks}
\begin{figure}[htp]
    \centering
    \begin{subfigure}{.5\textwidth}
        \centering
        \adjustbox{valign=c,frame}{\includegraphics[width=.9\linewidth]{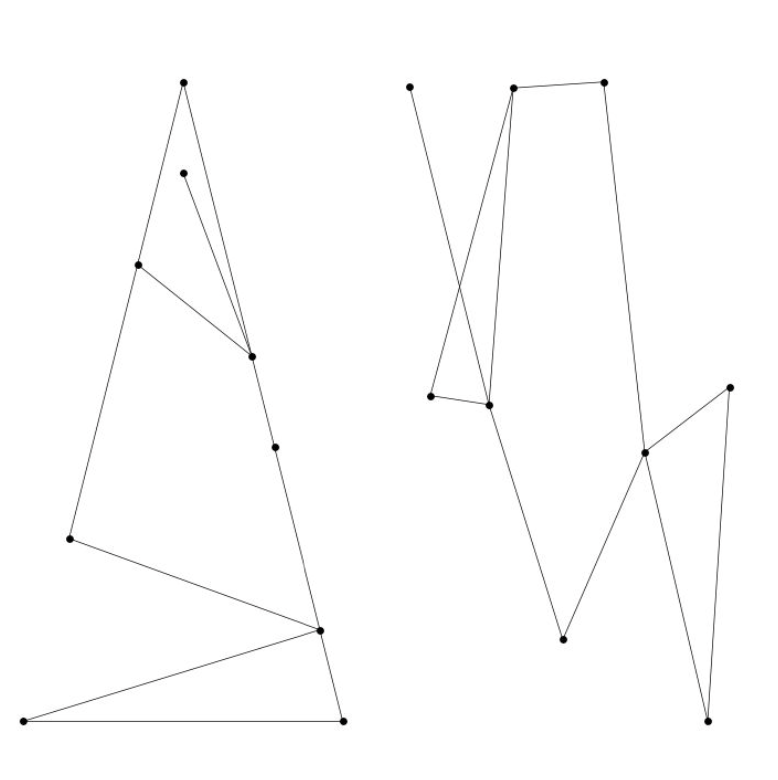}}
        \caption{True}
        \label{fig:sub1}
    \end{subfigure}%
    \begin{subfigure}{.5\textwidth}
        \centering
        \adjustbox{valign=c,frame}{\includegraphics[width=.9\linewidth]{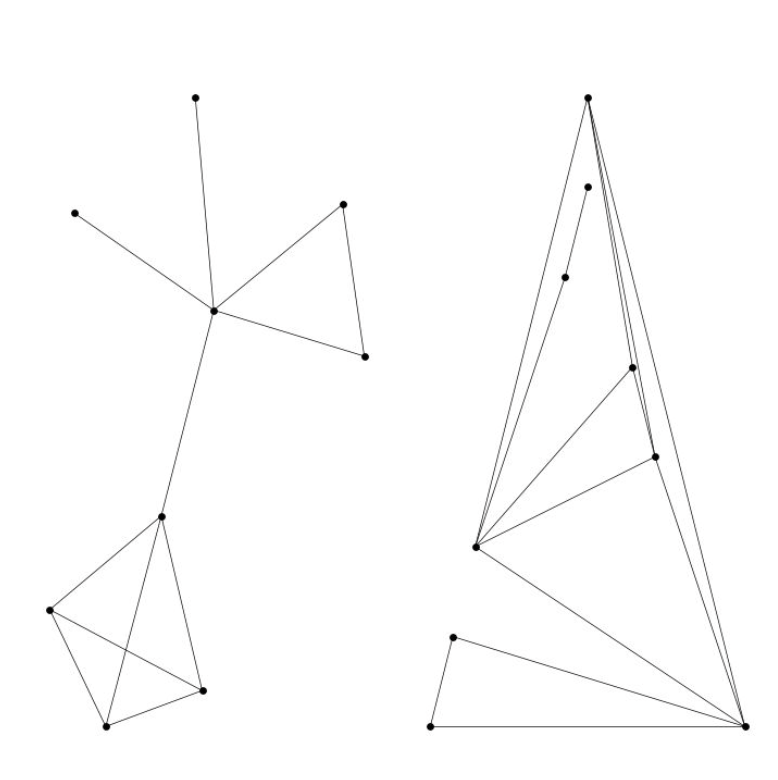}}
        \caption{False}
        \label{fig:sub2}
    \end{subfigure}
    \caption{Examples of the easy isomorphism task with true and false labels.}
    \label{fig:isomorphism_easy}
\end{figure}

\textbf{Easy Isomorphism:} For the easy isomorphism tasks, we plotted two graphs in a single plot, with the graph on the right using the Kamada-Kawai layout and the graph on the left using the planar layout. As not all graphs are planar, we first identified and selected all planar graphs from the set of connected graphs. We then split these planar graphs into separate test and train sets, ensuring that no graph from the training set appeared in the test set. The true and false samples were created as follows:
\begin{itemize}
\item \textbf{True Labels:} For each true label, we took a graph and plotted it using both the Kamada-Kawai and planar layouts.
\item \textbf{False Labels:} For each false label, we plotted one graph using the Kamada-Kawai layout and then plotted another randomly selected graph from the set using the planar layout.
\end{itemize}
To maintain a balanced dataset, we ensured an equal number of true and false labels in both the training and test sets.
Figure \ref{fig:isomorphism_easy} presents examples of the easy isomorphism task with both true and false labels.

\begin{figure}[htp]
    \centering
    \begin{subfigure}{.5\textwidth}
        \centering
        \adjustbox{valign=c,frame}{\includegraphics[width=.9\linewidth]{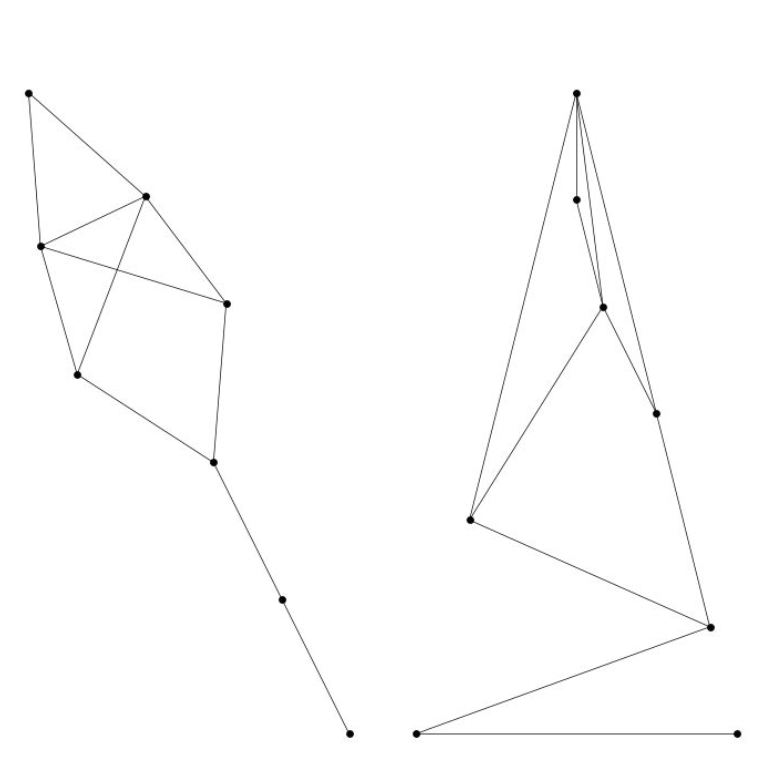}}
        \caption{True}
        \label{fig:sub1}
    \end{subfigure}%
    \begin{subfigure}{.5\textwidth}
        \centering
        \adjustbox{valign=c,frame}{\includegraphics[width=.9\linewidth]{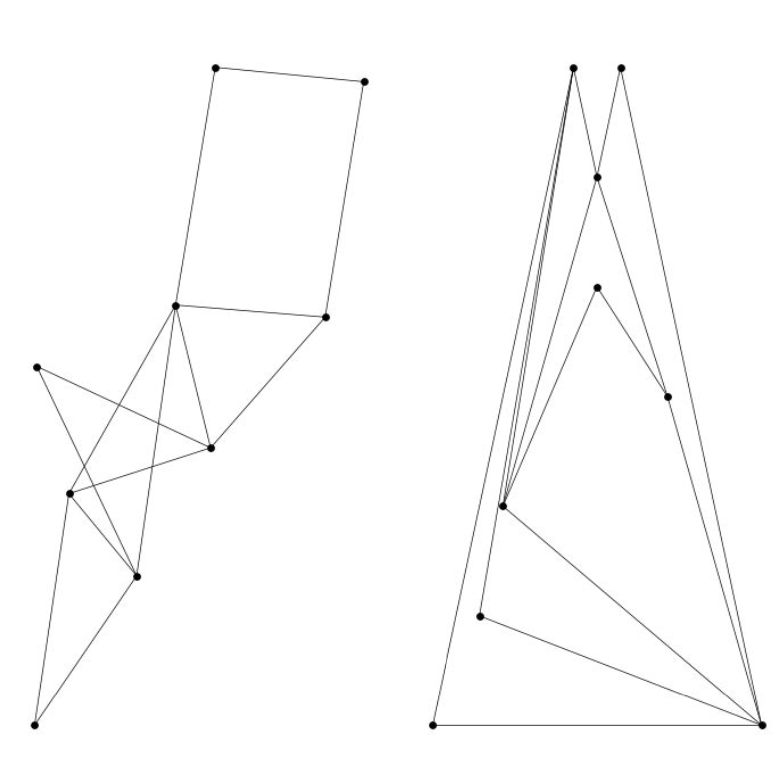}}
        \caption{False}
        \label{fig:sub2}
    \end{subfigure}
    \caption{Examples of the easy isomorphism task with true and false labels.}
    \label{fig:isomorphism_hard}
\end{figure}

\textbf{Hard Isomorphism:} For this task, we proceeded similarly to the previous one, but with a key difference in the selection of graphs for false labels. Instead of picking a random graph from the set, we searched for graphs within the set that are degree-equivalent to the first graph. From these degree-equivalent graphs, we randomly chose one and plotted it using the planar layout. In cases where no degree-equivalent graphs were found for the first graph within the set, we discarded that sample. As a result, the number of samples in this task is smaller compared to the easy isomorphism task. Figure \ref{fig:isomorphism_hard} presents examples of the hard isomorphism task with both true and false labels.

\subsection{Path Finding Tasks}
\begin{figure}[htp]
    \centering
    \begin{subfigure}{.5\textwidth}
        \centering
        \adjustbox{valign=c,frame}{\includegraphics[width=.9\linewidth]{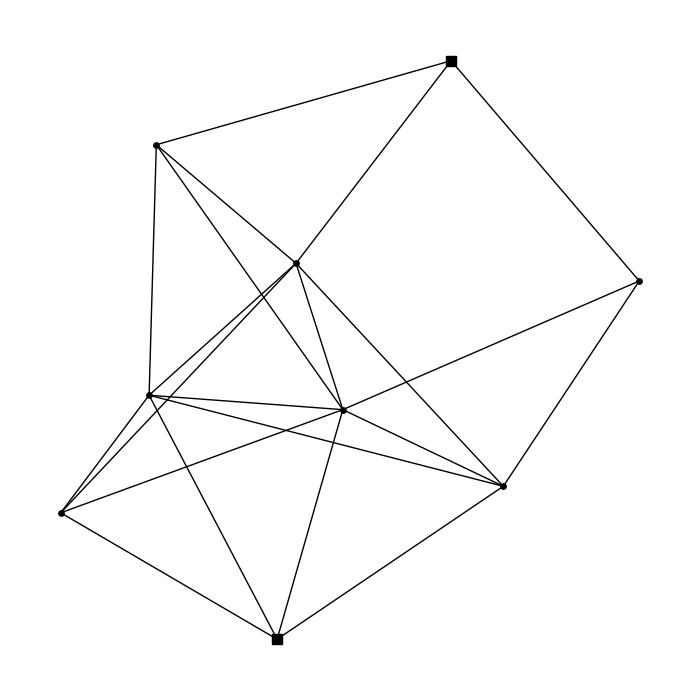}}
        \caption{The shortest path is of length 3.}
        \label{fig:sub1}
    \end{subfigure}%
    \begin{subfigure}{.5\textwidth}
        \centering
        \adjustbox{valign=c,frame}{\includegraphics[width=.9\linewidth]{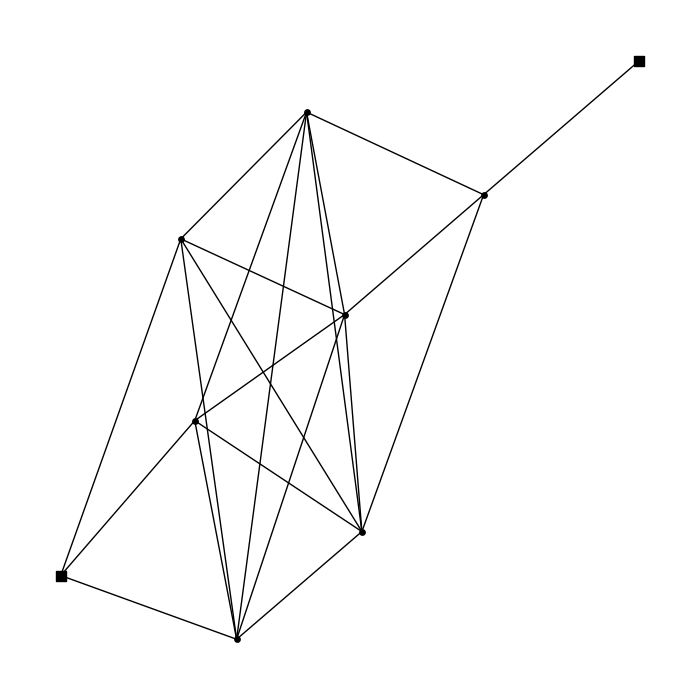}}
        \caption{The shortest path is of length 4.}
        \label{fig:sub2}
    \end{subfigure}
    \caption{Examples of the shortest path task between marked nodes.}
    \label{fig:shortestpath}
\end{figure}

\textbf{Shortest Path:}
For the shortest path task, we created a balanced set of graphs with shortest paths of lengths 1, 2, 3, and 4 between two marked nodes. We opted not to include graphs with shortest paths of length 5 or longer, as the number of such graphs within the 8-9 node graph set was too small to maintain a balanced dataset with an equal number of samples for each path length.
We began by separating the planar graphs for the test set. From the remaining non-planar graphs, which were used for the training set, we identified pairs of nodes with shortest paths of lengths 1, 2, 3, and 4. These nodes were marked with squares in the plot, while the rest of the nodes were represented as circles. The training set graphs were plotted using the Kamada-Kawai layout. For each path length, we plotted 20,000 samples, resulting in a total of 80,000 samples for the training set.
For the test set, we created a similarly balanced set of graphs with different path lengths and plotted them using the planar layout.
Figure \ref{fig:shortestpath} shows examples of the shortest path task.

\begin{figure}[htp]
    \centering
    \begin{subfigure}{.5\textwidth}
        \centering
        \adjustbox{valign=c,frame}{\includegraphics[width=.9\linewidth]{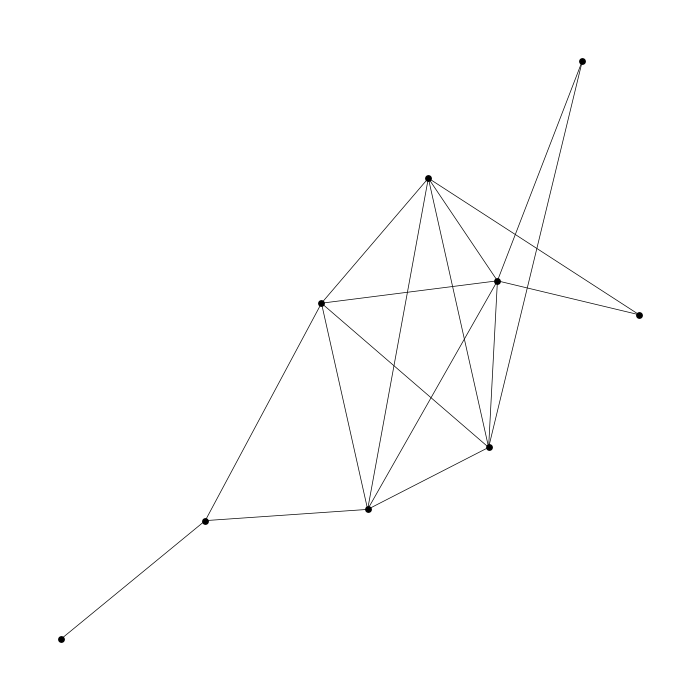}}
        \caption{True}
        \label{fig:sub1}
    \end{subfigure}%
    \begin{subfigure}{.5\textwidth}
        \centering
        \adjustbox{valign=c,frame}{\includegraphics[width=.9\linewidth]{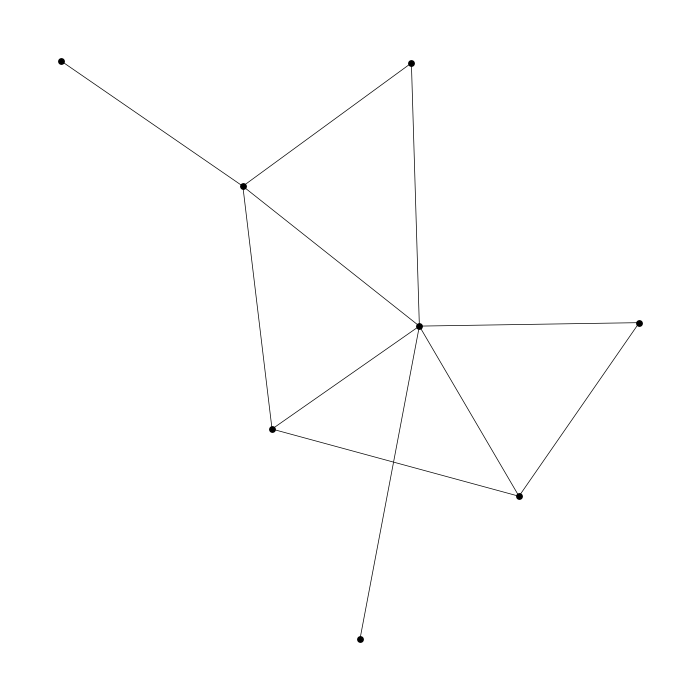}}
        \caption{False}
        \label{fig:sub2}
    \end{subfigure}
    \caption{Examples of the Hamiltonian path task with true and false labels.}
    \label{fig:path_hamiltonian}
\end{figure}

\textbf{Hamiltonian Paths:} For this task, we first identified the graphs that contain a Hamiltonian path from the entire set of 8-9 connected graphs. To create a balanced dataset with an equal number of true and false labels, we then randomly selected the same number of graphs from the remaining non-Hamiltonian graphs as the number of Hamiltonian graphs found. This step was necessary since Hamiltonian graphs are fewer in count compared to non-Hamiltonian graphs. After creating the balanced dataset, we split the set into train and test subsets. For the training set, we plotted the graphs using the Kamada-Kawai layout, while for the test set, we used the planar layout. 
Figure \ref{fig:path_hamiltonian} shows examples of the Hamiltonian path task.

\subsection{Cycle Finding Tasks}

\begin{figure}[htp]
    \centering
    \begin{subfigure}{.5\textwidth}
        \centering
        \adjustbox{valign=c,frame}{\includegraphics[width=.9\linewidth]{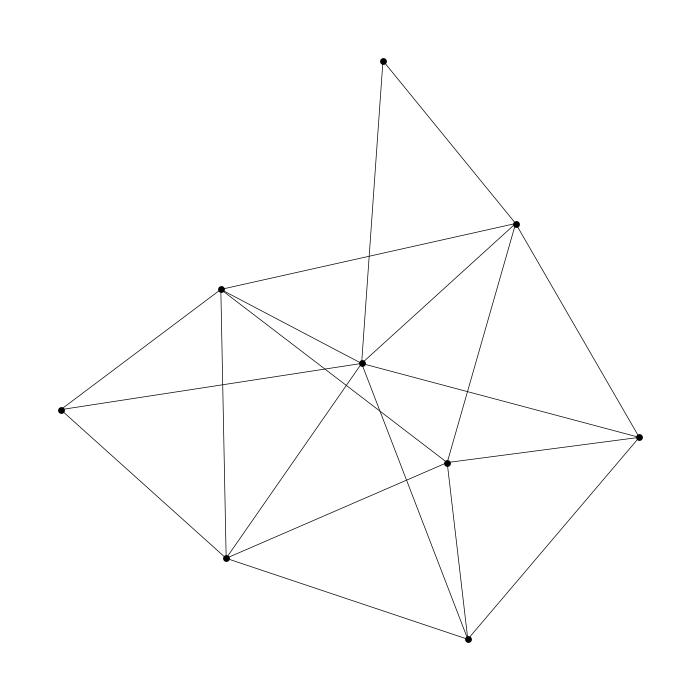}}
        \caption{True}
        \label{fig:sub1}
    \end{subfigure}%
    \begin{subfigure}{.5\textwidth}
        \centering
        \adjustbox{valign=c,frame}{\includegraphics[width=.9\linewidth]{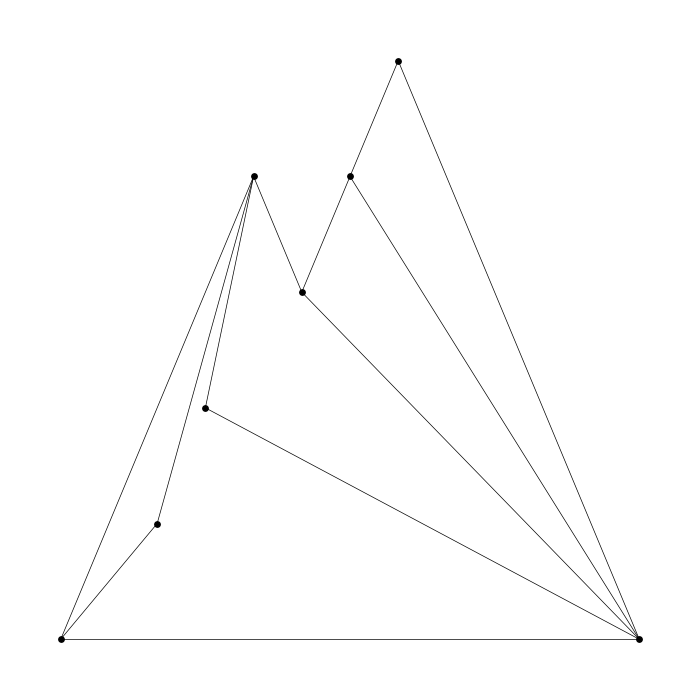}}
        \caption{False}
        \label{fig:sub2}
    \end{subfigure}
    \caption{Examples of the Hamiltonian cycle task with true and false labels.}
    \label{fig:cycle_hamiltonian}
\end{figure}

\textbf{Hamiltonian Cycles:} The task of finding Hamiltonian cycles was approached in a similar manner to the Hamiltonian path task. We first identified graphs that contain a Hamiltonian cycle from the entire set of 8-9 connected graphs. To create a balanced dataset with an equal number of true and false labels, we randomly selected the same number of graphs from the remaining non-Hamiltonian cycle graphs as the number of Hamiltonian cycle graphs found. After creating the balanced dataset, we split the set into train and test subsets. For the training set, we plotted the graphs using the Kamada-Kawai layout, while for the test set, we used the planar layout. Figure \ref{fig:cycle_hamiltonian} presents examples of the Hamiltonian cycle task.

\textbf{Chordless Cycles:} The chordless cycle task was similar to the shortest path task, but instead focused on the largest chordless cycle lengths of 3, 4, 5, and 6. We chose not to include graphs with no chordless cycles (trees) because the number of such graphs was too small to maintain a balanced set.

We first separated the planar graphs for the test set. From the remaining non-planar graphs, which were used for the training set, we identified the largest chordless cycles of lengths 3, 4, 5, and 6. For each chordless cycle length, we plotted an equal number of samples, resulting in a balanced dataset. The training set graphs were plotted using the Kamada-Kawai layout.

For the test set, we created a similarly balanced set of graphs with different chordless cycle lengths and plotted them using the planar layout.
Figure \ref{fig:chordless} shows examples of the chordless cycles task.

\section{Training the Vision Models}

For training the vision models on the Visual Graph Arena benchmark tasks, we utilized three state-of-the-art architectures: ConvNeXt base, Swin Transformer base, and ViT (Vision Transformer) base. All models were initially pre-trained on the ImageNet dataset and then fine-tuned on our specific graph-based tasks. 
The training process was implemented using PyTorch, with the Adam optimizer and a learning rate of $1e-4$. The batch size was set to 32, and the models were trained for 10 epochs on each task. The CrossEntropyLoss was used as the loss function, as the tasks were formulated as either binary or multi-class classification problems. We used 2 NVIDIA TITAN RTX GPUs for training the models.

The input images were preprocessed using transformations from the PyTorch \texttt{transforms} module. The images were resized to $384 \times 384$ pixels, and the pre-trained models were trained on the same size of images on ImageNet.

To handle the different output requirements of the tasks, the number of output classes in the final layer of the models was adjusted accordingly. For the isomorphism tasks and the Hamiltonian path/cycle tasks, the number of output classes was set to 2 (binary classification). For the shortest path and chordless cycle tasks, the number of output classes was set to 4, corresponding to the different path lengths or cycle sizes.

During training, the model's performance was evaluated on the test set at the end of each epoch. The test set consisted of graphs drawn using different layouts than those used in the training set, allowing us to assess the model's ability to generalize to different visual representations. After the training process, the best-performing model on the test set was selected for each task and architecture combination. The code for training the vision models, written in Python using PyTorch, is provided in the supplementary materials and will be publicly available in a GitHub repository. 

\begin{figure}[htp]
    \centering
    \begin{subfigure}{.5\textwidth}
        \centering
        \adjustbox{valign=c,frame}{\includegraphics[width=.9\linewidth]{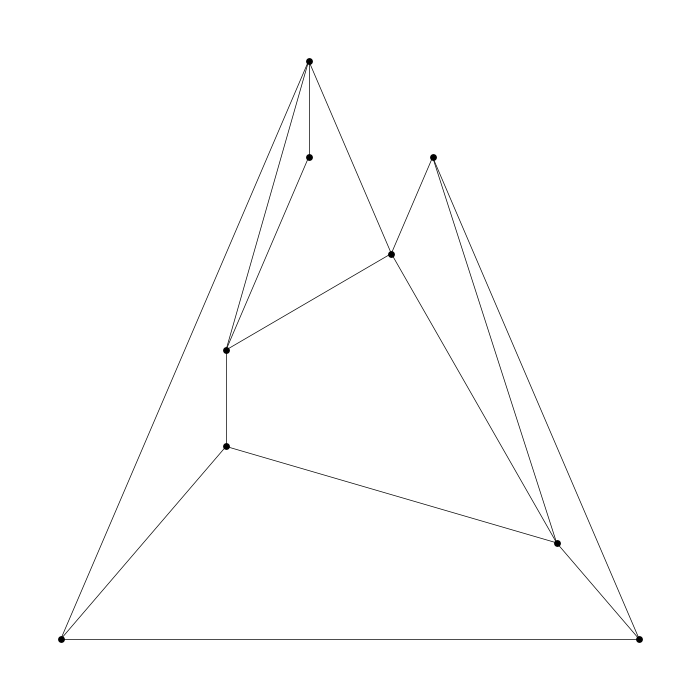}}
        \caption{The biggest chordless cycle is of length 6.}
        \label{fig:sub1}
    \end{subfigure}%
    \begin{subfigure}{.5\textwidth}
        \centering
        \adjustbox{valign=c,frame}{\includegraphics[width=.9\linewidth]{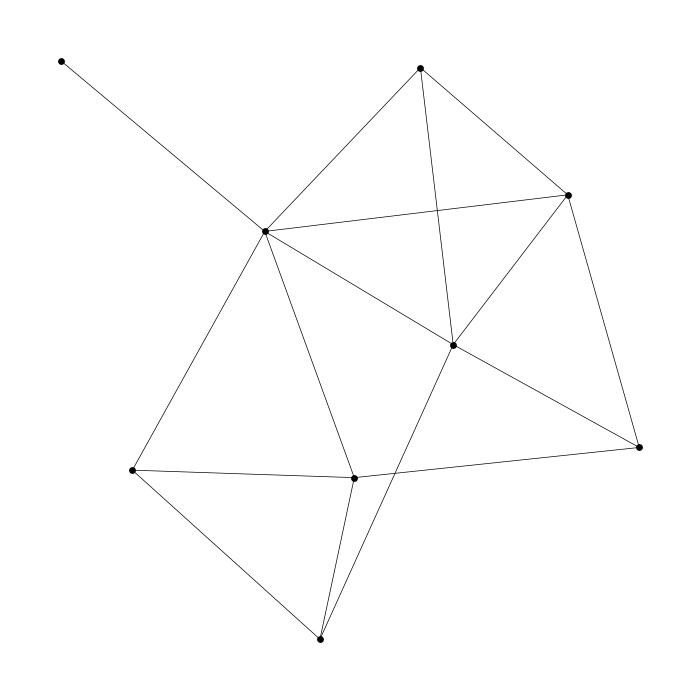}}
        \caption{The biggest chordless cycle is of length 4.}
        \label{fig:sub2}
    \end{subfigure}
    \caption{Examples of the biggest chordless cycle task.}
    \label{fig:chordless}
\end{figure}

\section{MLLMs adjacency matrix creation}

In this section, we evaluate the performance of GPT-4o and o1, Claude 3 Opus, and Claude 3.5 Sonnet in transcribing graph structures. We presented these models with three graph images and provided the prompt: "The image presented contains a graph. Please transcribe this graph's adjacency matrix." We then analyzed their ability to generate accurate adjacency matrices from the visual input. Our results reveal that all models struggle to correctly transcribe the graphs, with varying degrees of accuracy in extracting basic graph properties. GPT-4 comes close to accurately identifying the number of nodes but struggles with edge counts. Claude 3.5 Sonnet shows the best performance, closely approximating both the number of nodes and edges. In contrast, Claude 3 Opus performs poorly across all metrics. We visualized the graphs transcribed by these models to assess their accuracy. Although Claude 3.5 Sonnet was not available at the time of our initial submission, we have incorporated its responses into our analysis, because it demonstrates superior performance compared to both GPT-4 and Claude 3 Opus, showing a marked improvement over Opus in particular.

\promptbox{The image presented contains a graph. Please transcribe this graph's adjacency matrix.}

\begin{figure}[htp]
    \centering
    \begin{subfigure}{.33\textwidth}
        \centering
        \adjustbox{valign=c,frame}{\includegraphics[width=.9\linewidth]{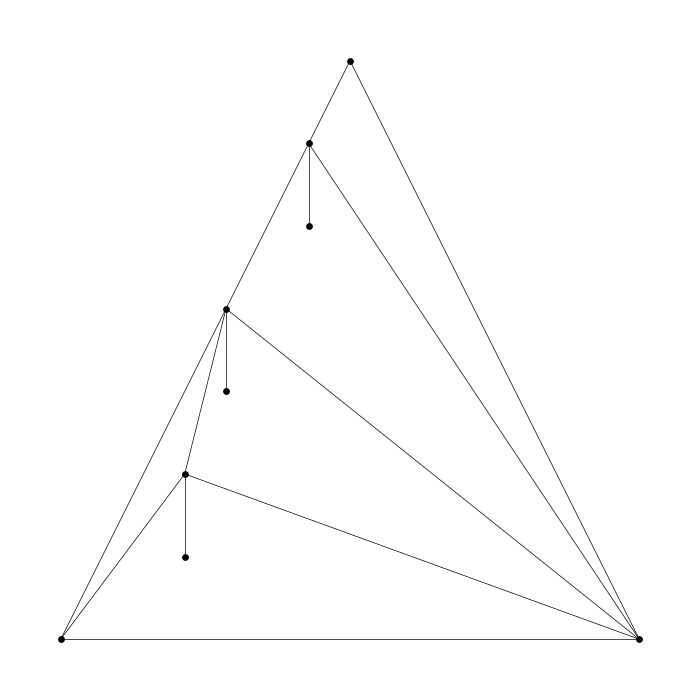}}
        \caption{Graph a}
        \label{fig:sub1}
    \end{subfigure}%
    \begin{subfigure}{.33\textwidth}
        \centering
        \adjustbox{valign=c,frame}{\includegraphics[width=.9\linewidth]{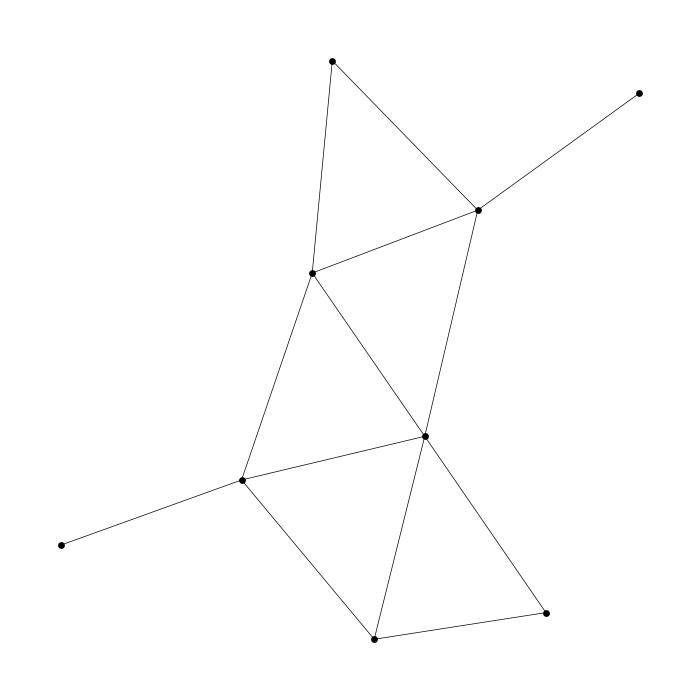}}
        \caption{Graph b}
        \label{fig:sub2}
    \end{subfigure}
    \begin{subfigure}{.33\textwidth}
        \centering
        \adjustbox{valign=c,frame}{\includegraphics[width=.9\linewidth]{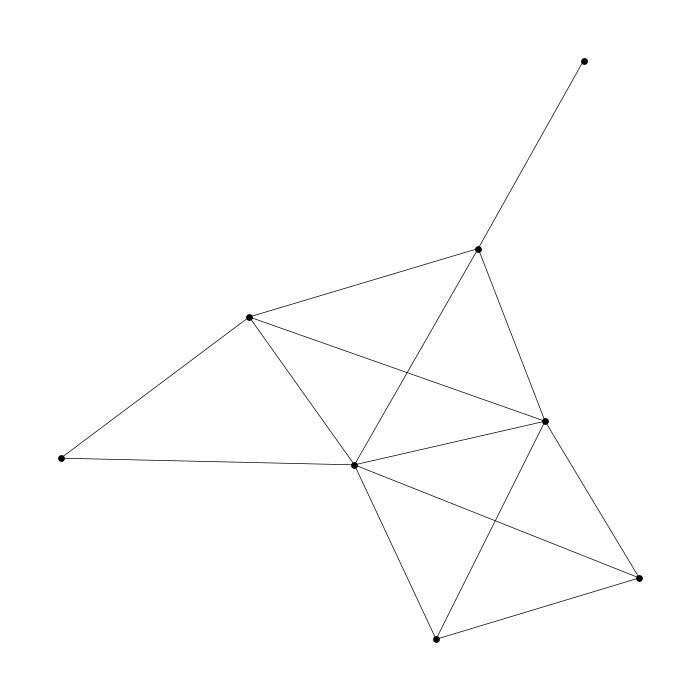}}
        \caption{Graph c}
        \label{fig:sub2}
    \end{subfigure}
    \caption{The three graphs we asked the LLMs to generate the adjacency matrix.}
    \label{fig:chordless}
\end{figure}

\begin{figure}[h]
    \centering
    \begin{subfigure}{.33\textwidth}
        \centering
        \adjustbox{valign=c,frame}{\includegraphics[width=.9\linewidth]{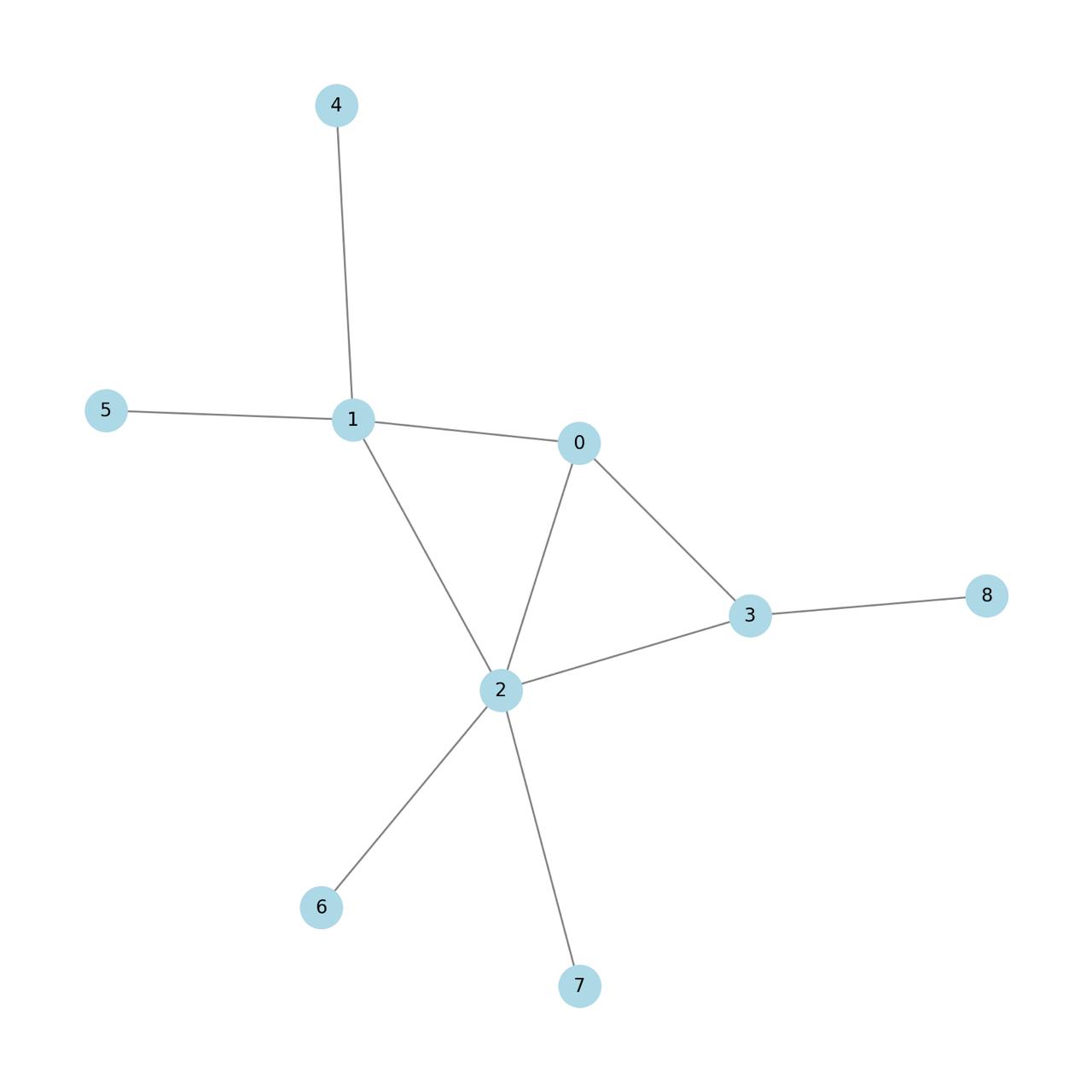}}
        \caption{Graph a}
        \label{fig:sub1}
    \end{subfigure}%
    \begin{subfigure}{.33\textwidth}
        \centering
        \adjustbox{valign=c,frame}{\includegraphics[width=.9\linewidth]{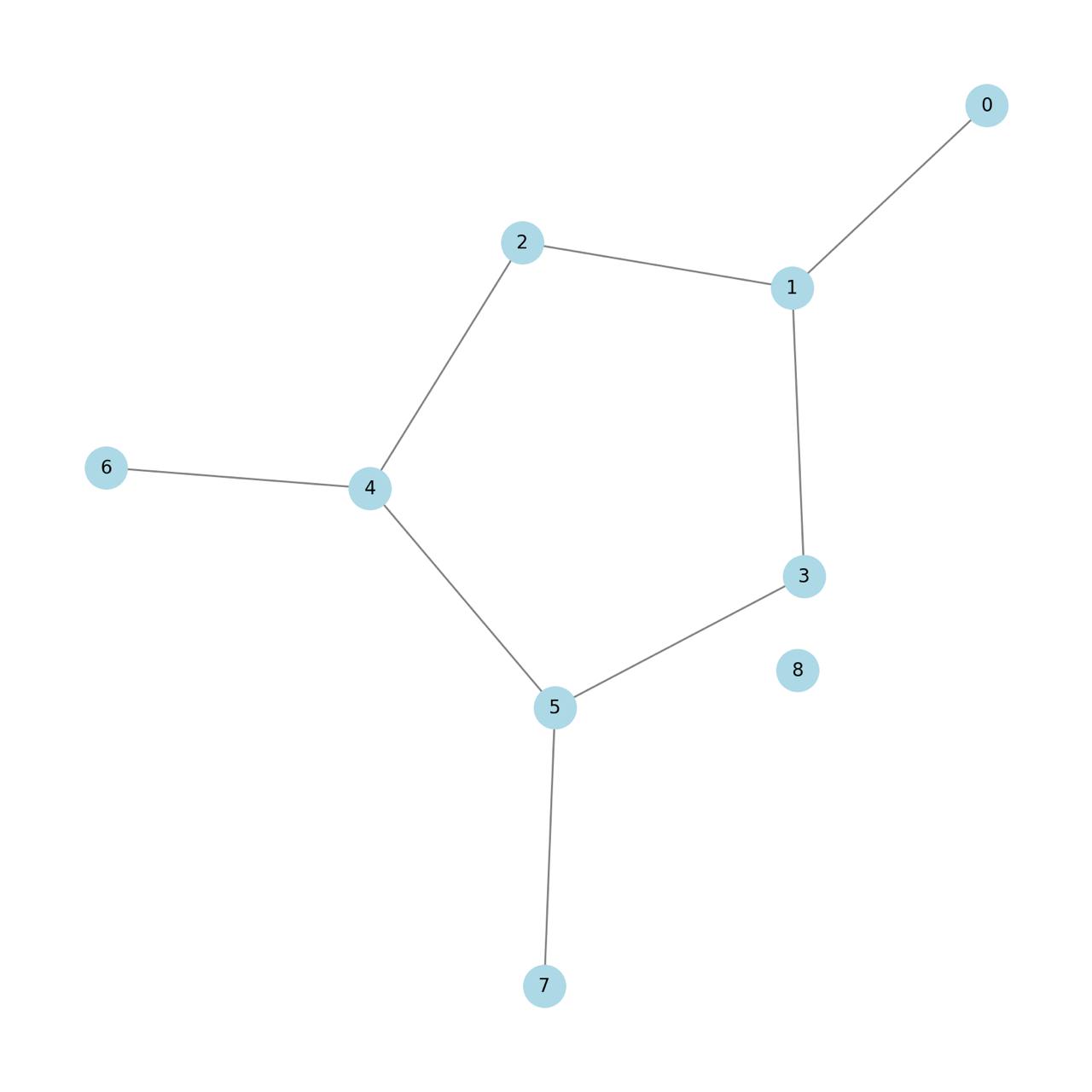}}
        \caption{Graph b}
        \label{fig:sub2}
    \end{subfigure}
    \begin{subfigure}{.33\textwidth}
        \centering
        \adjustbox{valign=c,frame}{\includegraphics[width=.9\linewidth]{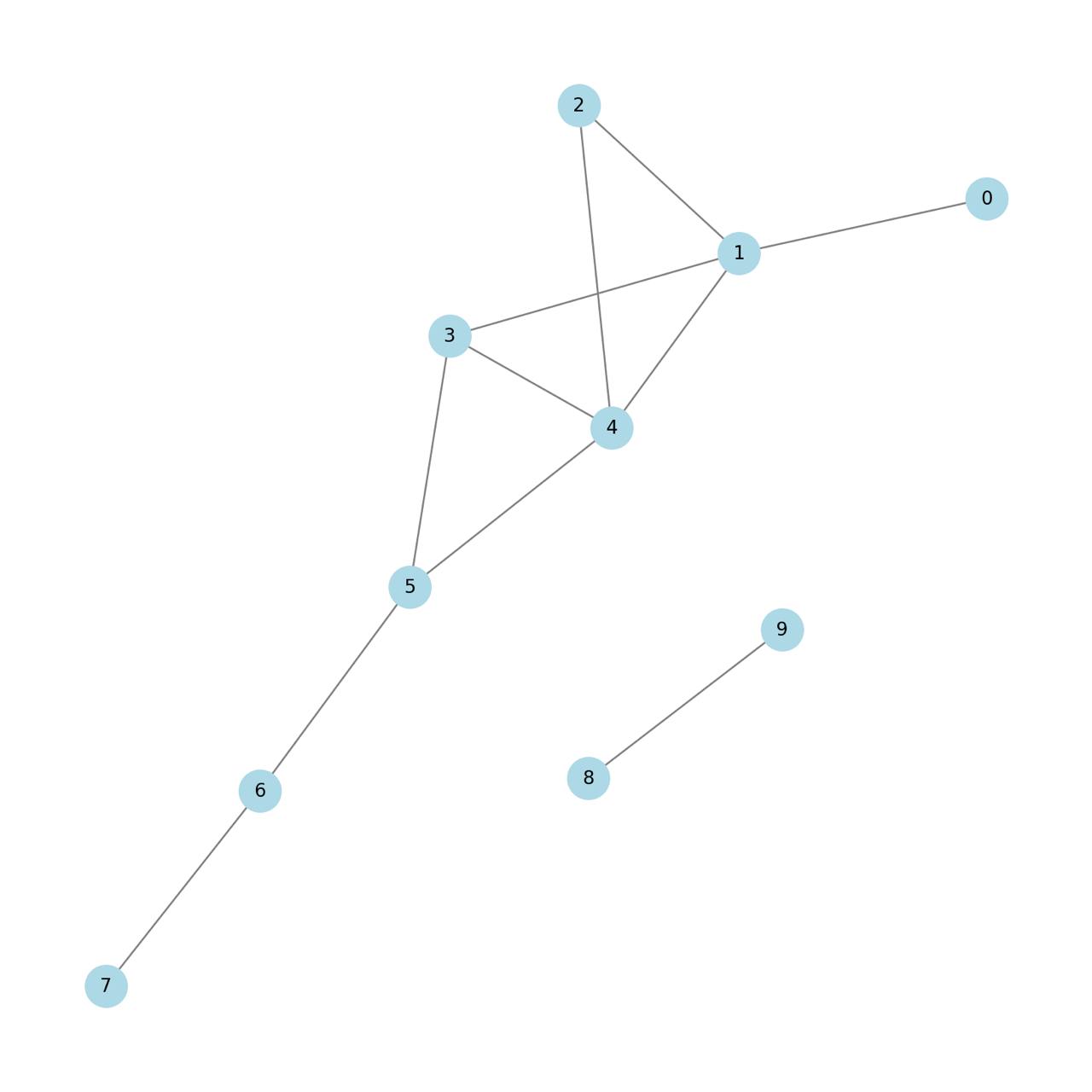}}
        \caption{Graph c}
        \label{fig:sub2}
    \end{subfigure}
    \caption{reconstructed graphs from the generated adjacency matrix by GPT-4o.}
    \label{fig:chordless}
\end{figure}

\begin{figure}[h]
    \centering
    \begin{subfigure}{.33\textwidth}
        \centering
        \adjustbox{valign=c,frame}{\includegraphics[width=.9\linewidth]{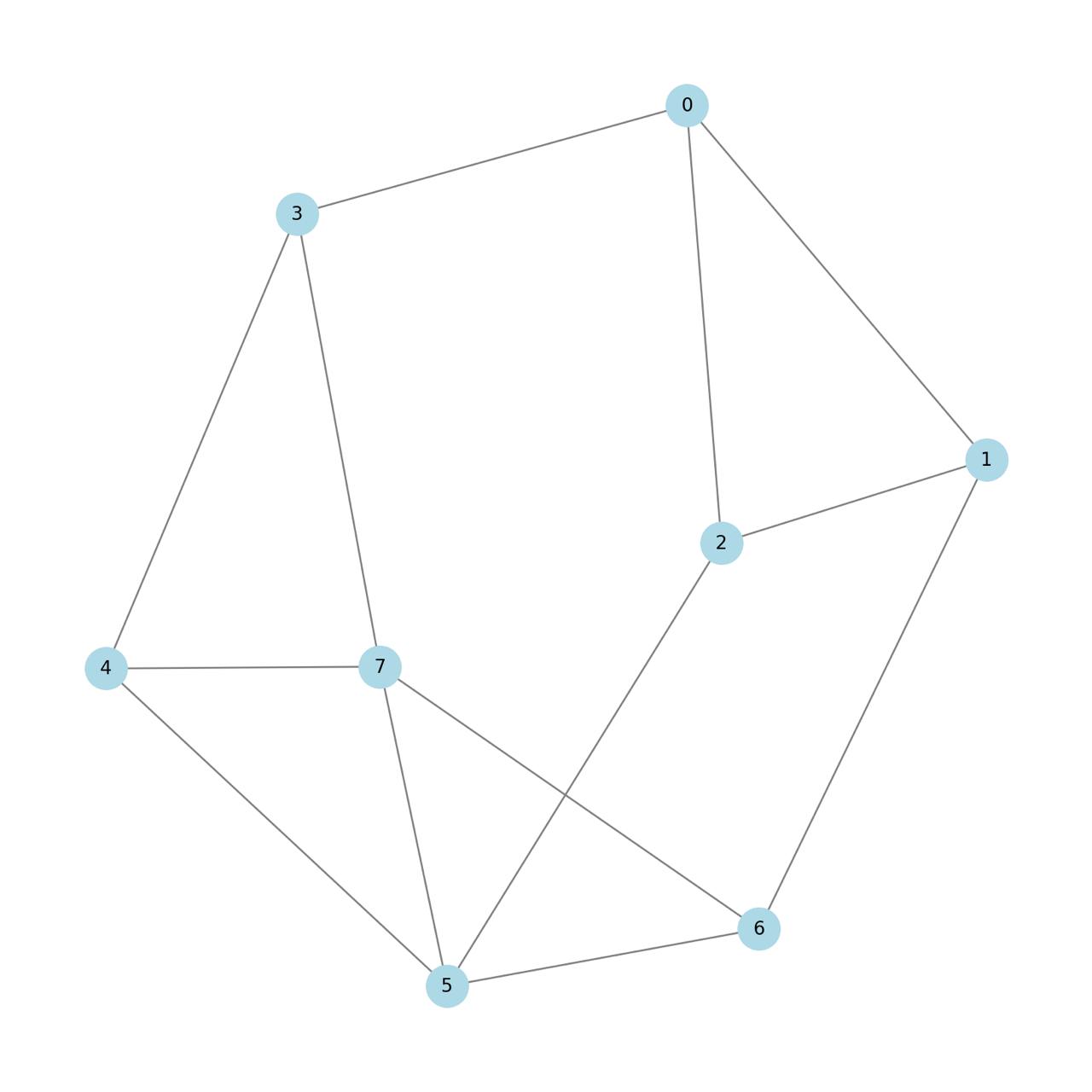}}
        \caption{Graph a}
        \label{fig:sub1}
    \end{subfigure}%
    \begin{subfigure}{.33\textwidth}
        \centering
        \adjustbox{valign=c,frame}{\includegraphics[width=.9\linewidth]{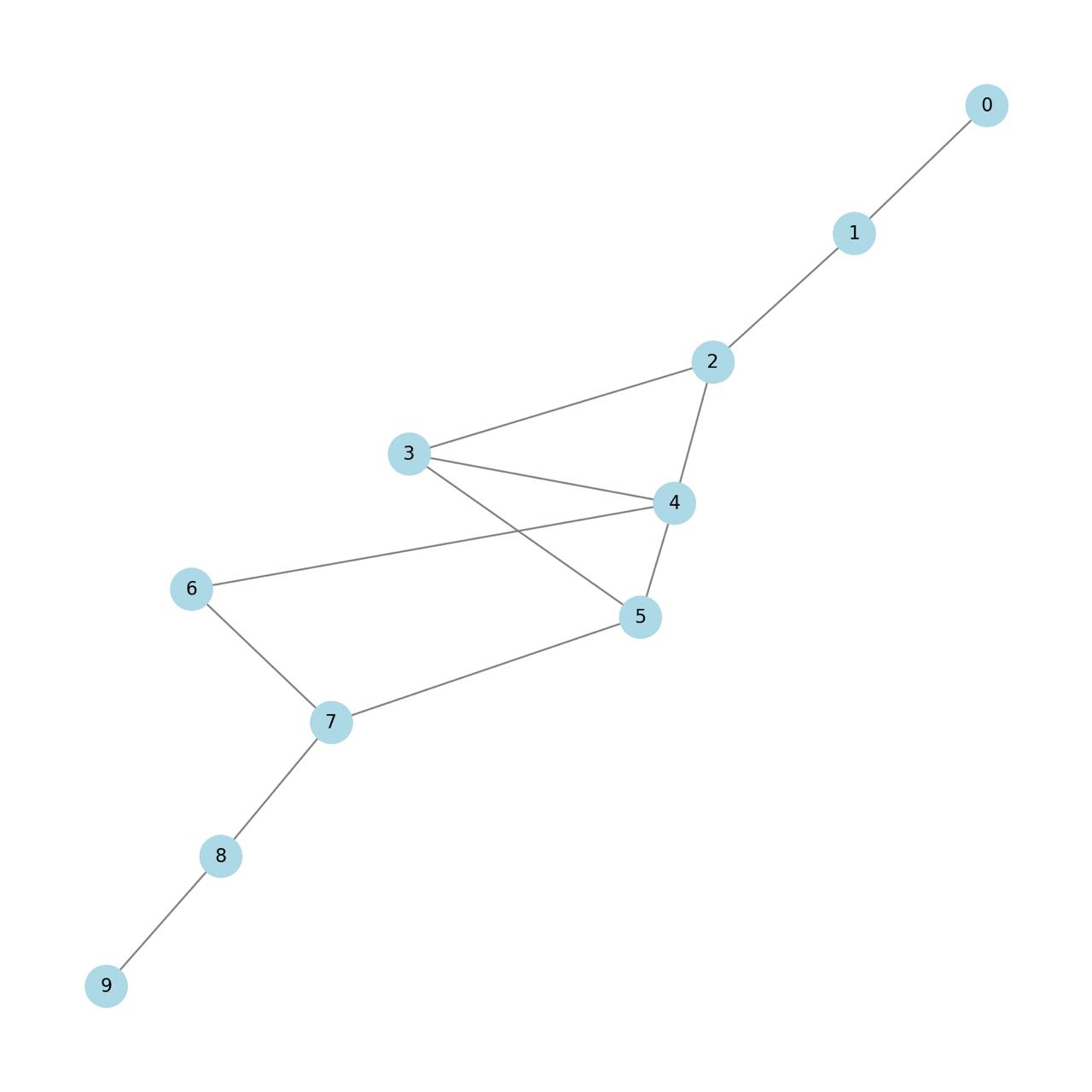}}
        \caption{Graph b}
        \label{fig:sub2}
    \end{subfigure}
    \begin{subfigure}{.33\textwidth}
        \centering
        \adjustbox{valign=c,frame}{\includegraphics[width=.9\linewidth]{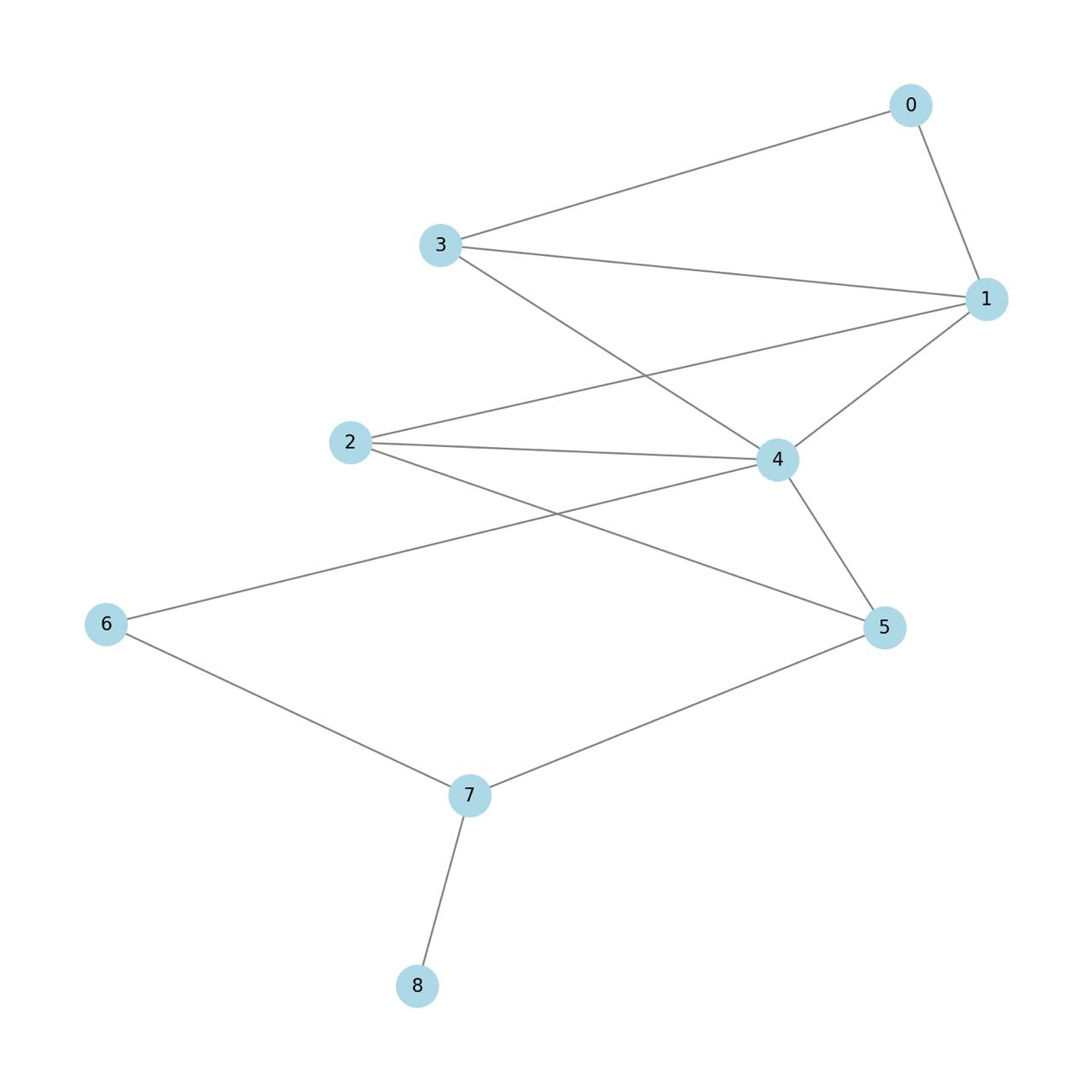}}
        \caption{Graph c}
        \label{fig:sub2}
    \end{subfigure}
    \caption{reconstructed graphs from the generated adjacency matrix by Claude 3.5 Sonnet.}
    \label{fig:chordless}
\end{figure}

\begin{figure}[h]
    \centering
    \begin{subfigure}{.33\textwidth}
        \centering
        \adjustbox{valign=c,frame}{\includegraphics[width=.9\linewidth]{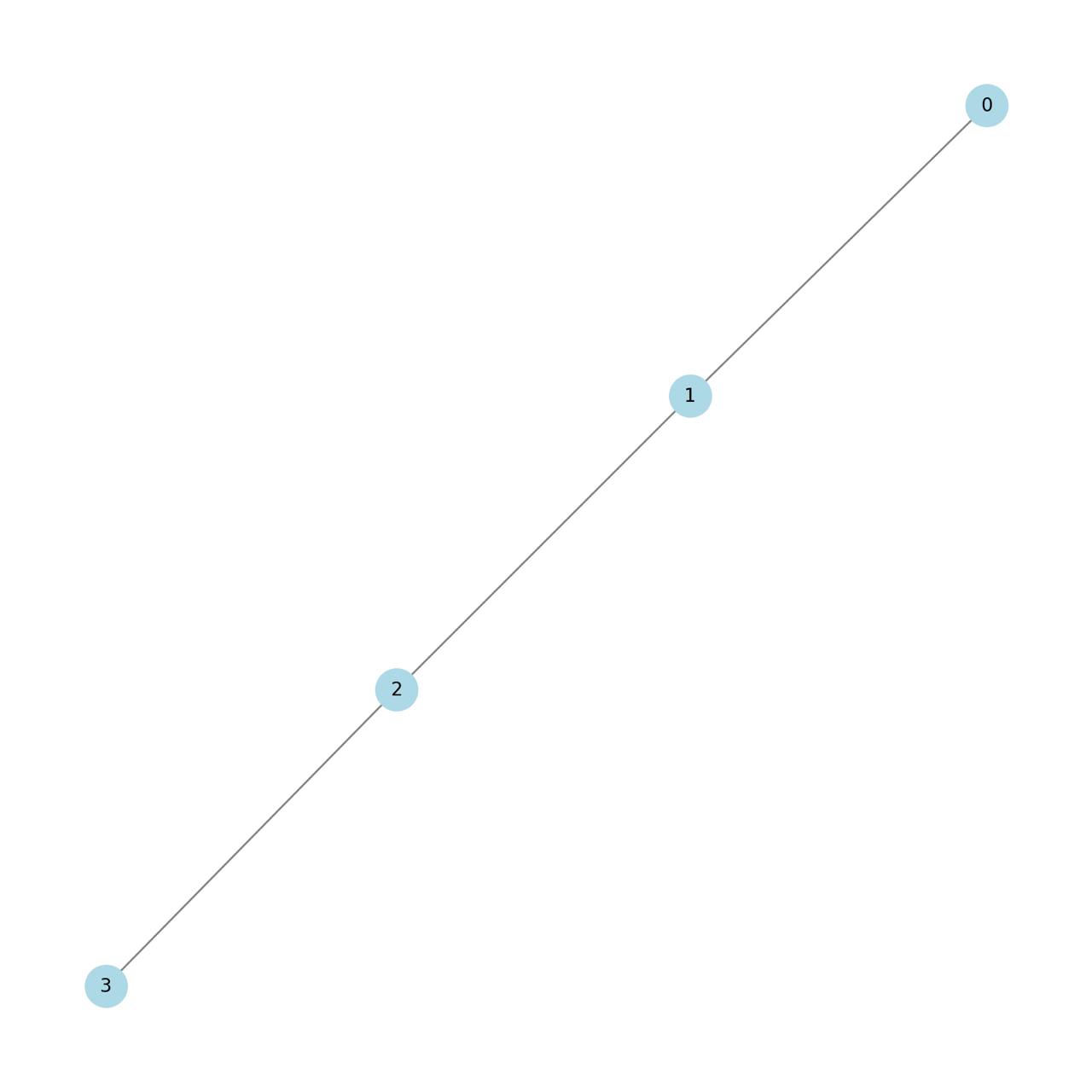}}
        \caption{Graph a}
        \label{fig:sub1}
    \end{subfigure}%
    \begin{subfigure}{.33\textwidth}
        \centering
        \adjustbox{valign=c,frame}{\includegraphics[width=.9\linewidth]{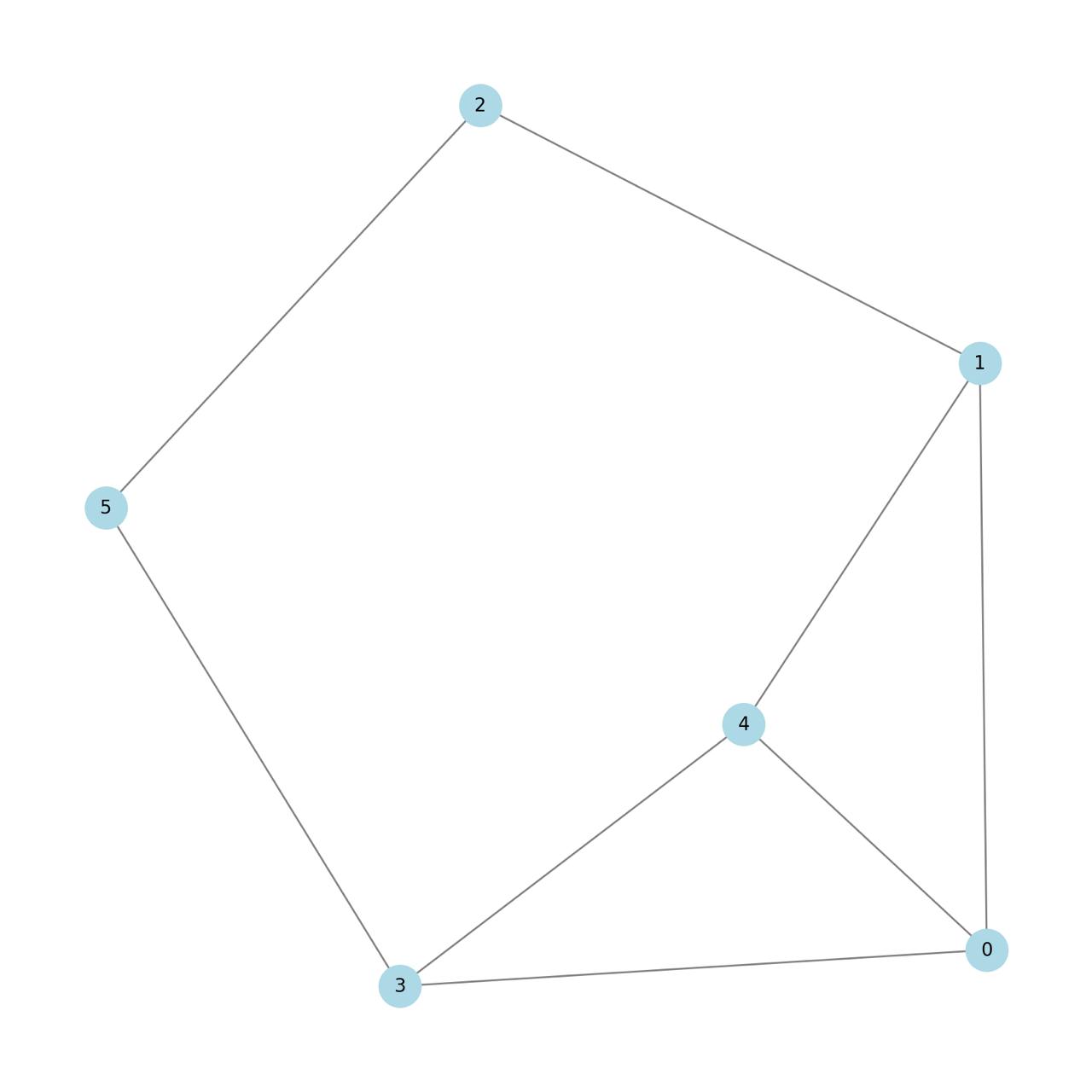}}
        \caption{Graph b}
        \label{fig:sub2}
    \end{subfigure}
    \begin{subfigure}{.33\textwidth}
        \centering
        \adjustbox{valign=c,frame}{\includegraphics[width=.9\linewidth]{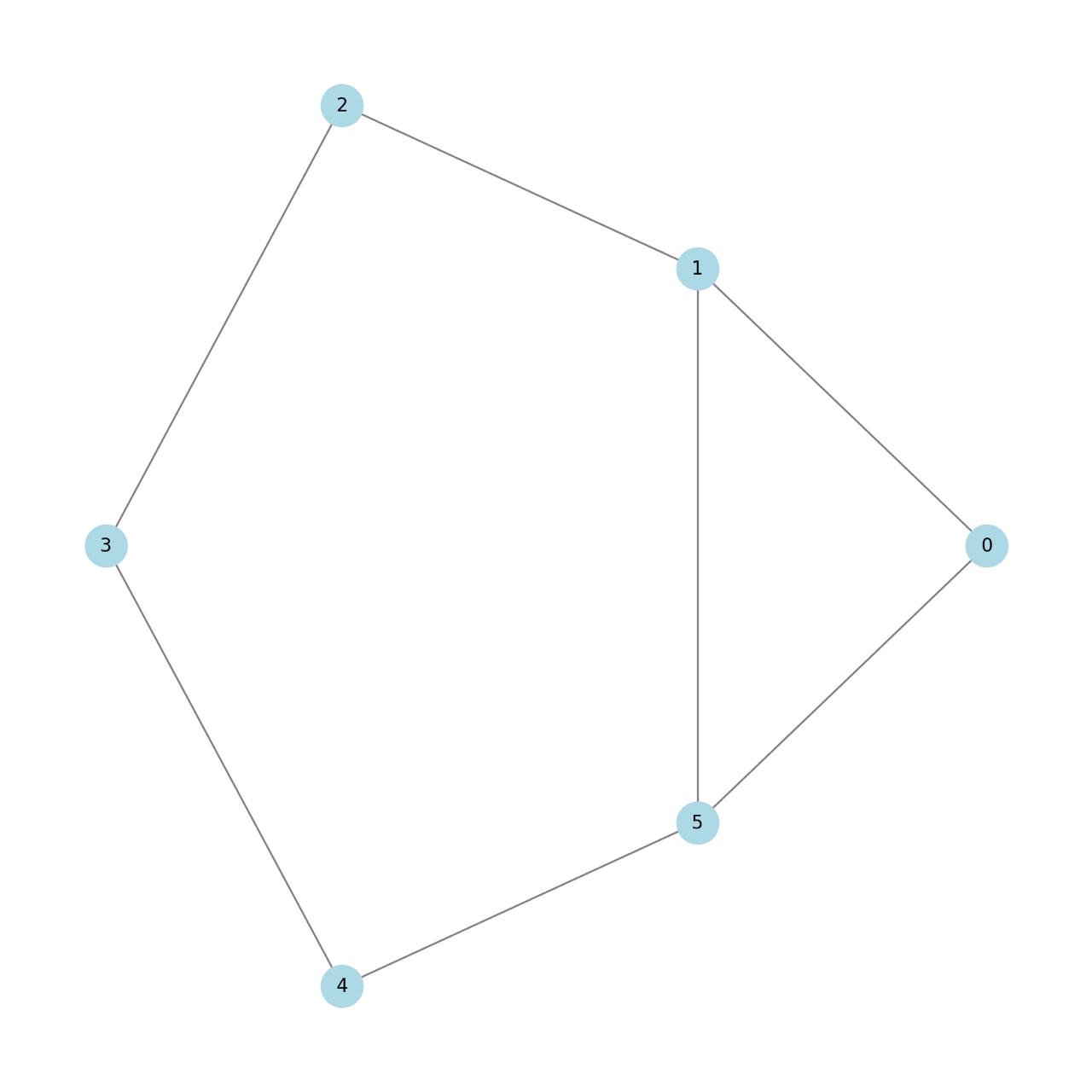}}
        \caption{Graph c}
        \label{fig:sub2}
    \end{subfigure}
    \caption{reconstructed graphs from the generated adjacency matrix by Claude 3 Opus.}
    \label{fig:chordless}
\end{figure}

\begin{figure}[h]
    \centering
    \begin{subfigure}{.33\textwidth}
        \centering
        \adjustbox{valign=c,frame}{\includegraphics[width=.9\linewidth]{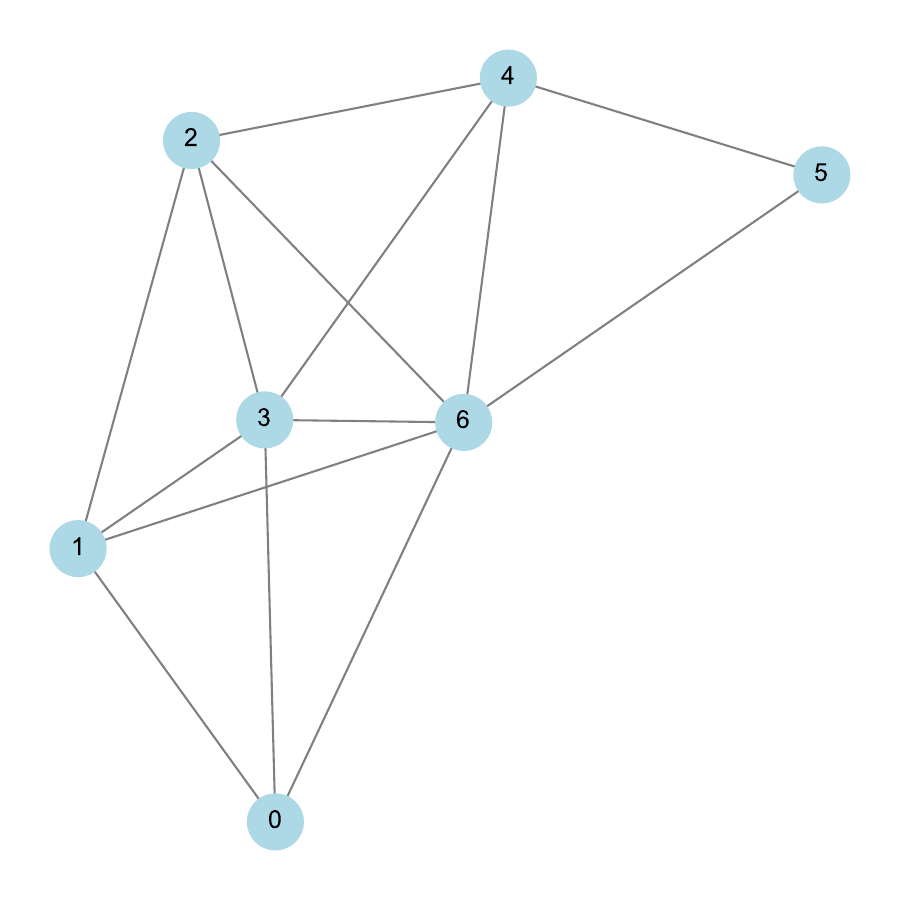}}
        \caption{Graph a}
        \label{fig:sub1}
    \end{subfigure}%
    \begin{subfigure}{.33\textwidth}
        \centering
        \adjustbox{valign=c,frame}{\includegraphics[width=.9\linewidth]{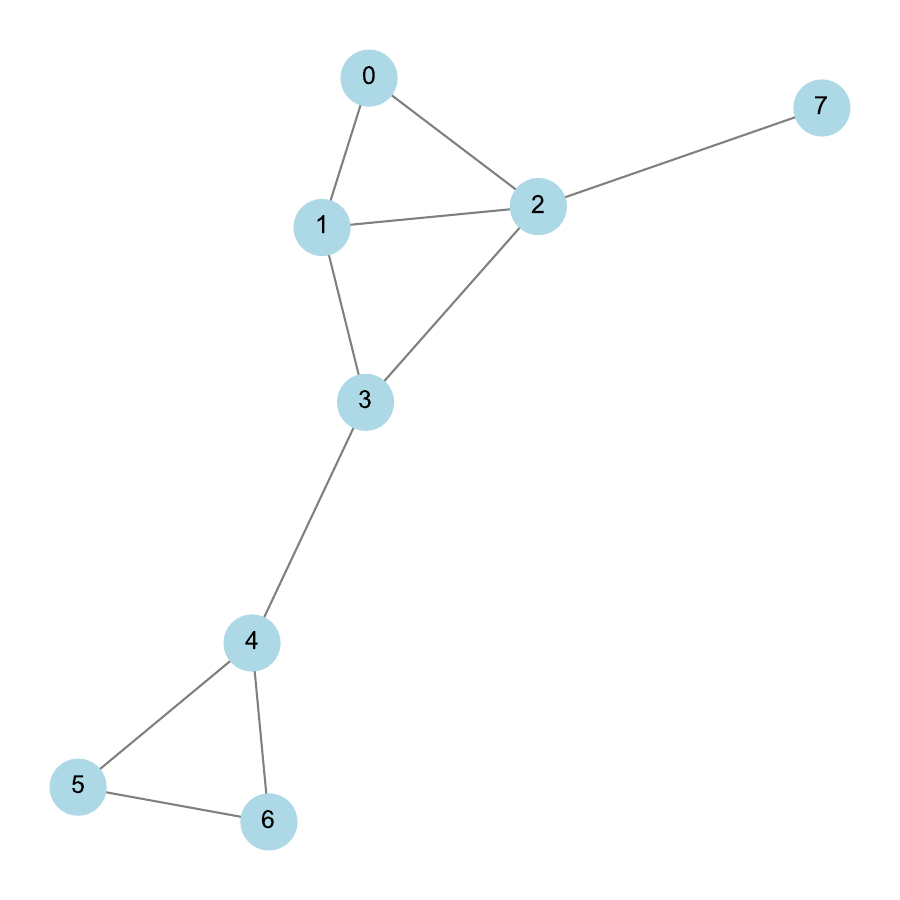}}
        \caption{Graph b}
        \label{fig:sub2}
    \end{subfigure}
    \begin{subfigure}{.33\textwidth}
        \centering
        \adjustbox{valign=c,frame}{\includegraphics[width=.9\linewidth]{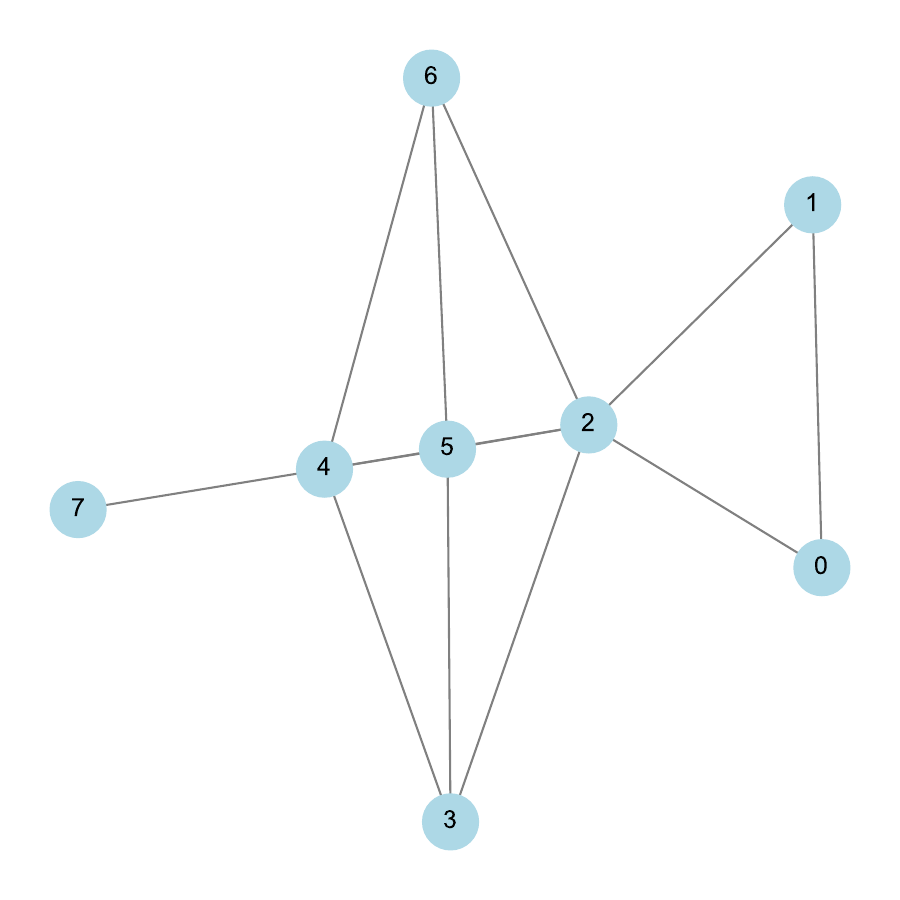}}
        \caption{Graph c}
        \label{fig:sub2}
    \end{subfigure}
    \caption{reconstructed graphs from the generated adjacency matrix by GPT o1.}
    \label{fig:chordless}
\end{figure}

\section{LLMs Prompts and Responses Samples}

In this section, we present one example of each task, along with the prompts used and the responses from GPT-o1, GPT-4, Claude 3 Opus, and Claude 3.5 Sonnet. We highlight correct final answers in green and incorrect ones in red. The results demonstrate that these models can analyze graphs, provide chains of thought, and attempt to track nodes and edges. However, they often fail to solve the problems accurately. In the example in three cases, while the final yes/no responses are correct, closer inspection of the analysis reveals obvious mistakes, such as incorrectly counting the number of nodes. The correct responses are not surprising given the yes/no nature of some tasks. Notably, in the shortest-path task, all models unexpectedly fail to detect that the two marked nodes are connected by a single edge.

\subsection{Isomorphism - Easy}

\promptbox{The image presented contains two side-by-side graphs. Without using code, please answer whether the two given graphs are isomorphic. Please report your final answer as 'Yes' or 'No' as your very last word.}

\begin{figure}[h]
    \centering
    \fbox{\includegraphics[width=0.60\textwidth]{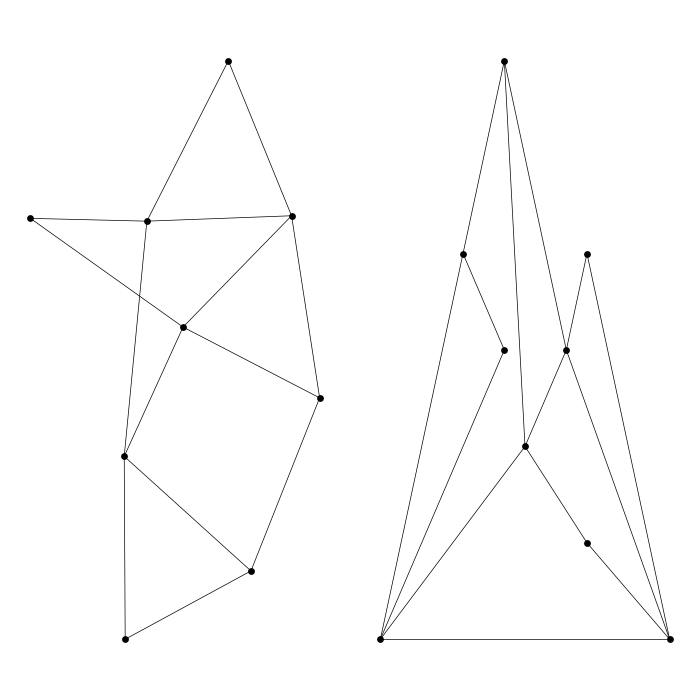}}
    \caption{The input image, the answer is 'Yes'}
    \label{fig:LLMresponse-easy}
\end{figure}

\gptfourbox{{To determine if the two graphs are isomorphic, we need to check whether there is a one-to-one correspondence between the vertices of the two graphs that preserves the adjacency relationships.}

{Looking at the two graphs in the image:}

\begin{enumerate}
    \item \textbf{Number of vertices}: Both graphs have the same number of vertices.
    \item \textbf{Number of edges}: Both graphs also have the same number of edges.
    \item \textbf{Vertex degrees}: The degrees of corresponding vertices (number of edges connected to each vertex) should be the same.
\end{enumerate}

{However, upon visual inspection, the structure of the graphs seems to differ significantly. The graph on the right has a more symmetrical structure, while the graph on the left appears more irregular. Specifically, the arrangement of connections and the overall shape of the graphs suggest that their connectivity patterns differ. This implies that there isn't a one-to-one correspondence that preserves the adjacency relationships between the vertices of the two graphs.}

{Therefore, the final answer is \colorbox{pink}{\textbf{No}}.}
}

\claudesonnetbox{To determine if these graphs are isomorphic, I'll analyze their structures:
Left graph:
\begin{itemize}
\item 9 vertices
\item 10 edges
\item Degree sequence: 3, 3, 3, 2, 2, 2, 2, 2, 1
\end{itemize}
Right graph:
\begin{itemize}
\item 9 vertices
\item 10 edges
\item Degree sequence: 4, 3, 2, 2, 2, 2, 2, 2, 1
\end{itemize}
While both graphs have the same number of vertices and edges, their degree sequences are different. The left graph has a maximum degree of 3, while the right graph has a vertex with degree 4.

Isomorphic graphs must have identical degree sequences. Since these graphs have different degree sequences, they cannot be isomorphic.

\colorbox{pink}{No}}
\claudeopusbox{To determine if the two graphs are isomorphic, I'll check if there exists a bijective mapping between their vertex sets that preserves adjacency.

Both graphs have 5 vertices. Let's label the vertices of the left graph L1 to L5 from top to bottom, and the vertices of the right graph R1 to R5 from top to bottom.

Comparing vertex degrees:

L1, L2, L4, L5 have degree 3. L3 has degree 4.

R1, R3, R5 have degree 3. R2, R4 have degree 4.

Since the degree sequences don't match, there cannot be an adjacency-preserving bijection between the vertex sets.

Therefore, the two graphs are not isomorphic. 

\colorbox{pink}{No}.}

\gptoOnebox{Thought about isomorphic graphs for 3m 21s

A quick way to see they cannot match is by comparing vertex‐degrees. In the right‐hand graph, the “apex” vertex (the topmost point) clearly has degree 5 (it has edges to five other vertices). In the left‐hand graph, no vertex has that many incident edges (each has degree at most 3 or 4). Since an isomorphism would require a one‐to‐one match of vertices with the same degree, the two graphs cannot be isomorphic.

\colorbox{pink}{No}}

\subsection{Isomorphism - Hard}

\promptbox{The image presented contains two side-by-side graphs. Without using code, please answer whether the two given graphs are isomorphic. Please report your final answer as 'Yes' or 'No' as your very last word.}

\begin{figure}[h]
    \centering
    \fbox{\includegraphics[width=0.60\textwidth]{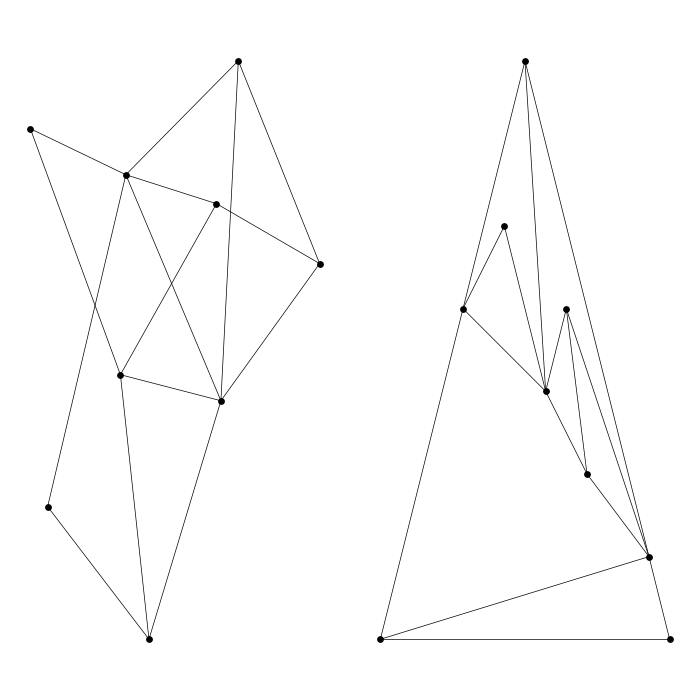}}
    \caption{The input image, the answer is 'No'}
    \label{fig:LLMresponse-hard}
\end{figure}

\gptfourbox{{To determine if the two graphs are isomorphic, I will analyze the structural properties of each graph, such as the number of vertices, edges, and the degree of each vertex, and compare them.}

\textbf{Step 1: Count the vertices and edges.}
\begin{itemize}
    \item \textbf{Left graph:}
    \begin{itemize}
        \item Number of vertices: 10
        \item Number of edges: 13
    \end{itemize}
    \item \textbf{Right graph:}
    \begin{itemize}
        \item Number of vertices: 10
        \item Number of edges: 13
    \end{itemize}
\end{itemize}

\textbf{Step 2: Compare the degree of each vertex.}
\begin{itemize}
    \item Both graphs should have the same degree sequence if they are isomorphic. The degree of each vertex is the number of edges connected to it.
\end{itemize}

\textbf{Step 3: Visual inspection.}
\begin{itemize}
    \item Compare the overall structure and connectivity to see if a one-to-one correspondence exists between the vertices and edges of the two graphs.
\end{itemize}

Upon close inspection, I can conclude that both graphs have the same number of vertices and edges, and they appear to have a similar degree sequence and connectivity structure. Given these observations, the graphs seem to be isomorphic.

\textbf{Final answer: \colorbox{pink}{Yes}.}
}
\claudesonnetbox{To determine if the two graphs are isomorphic, I'll analyze their structures:

Left graph:
\begin{itemize}
    \item \colorbox{pink}{11 vertices}
    \item 15 edges
    \item Complex structure with multiple triangles and intersecting lines
\end{itemize}

Right graph:
\begin{itemize}
    \item \colorbox{pink}{11 vertices}
    \item 15 edges
    \item More elongated structure, fewer triangles, more ``tree-like'' in parts
\end{itemize}

While both graphs have the same number of vertices and edges, their structures appear quite different:

\begin{enumerate}
    \item The left graph is more densely interconnected, with multiple triangles and a more compact shape.
    \item The right graph has a more stretched out, less symmetrical structure.
    \item The degree distribution (number of edges connected to each vertex) seems different between the two graphs.
    \item The left graph appears to have more cycles than the right graph.
\end{enumerate}

These structural differences suggest that there is no one-to-one mapping that could transform one graph into the other while preserving all connections.

Given these observations, my conclusion is that these graphs are not isomorphic.

\colorbox{green}{\textbf{No}}}
\claudeopusbox{Okay, let's carefully analyze the two graphs to determine if they are isomorphic.
To be isomorphic, the two graphs must have:
\begin{enumerate}
\item The same number of vertices
\item The same number of edges
\item Equivalent connectivity between corresponding pairs of vertices
\end{enumerate}
Comparing the two given graphs:
\begin{enumerate}
\item Number of vertices: \
Both graphs appear to have 6 vertices each. So this condition is satisfied.
\item Number of edges: \
I count 9 edges in the graph on the left, and 9 edges in the graph on the right as well. Edge count matches.
\item Vertex connectivity: \
This is trickier to verify visually. We'd need to find a one-to-one mapping between the vertices of the two graphs, such that the edge connections are preserved.
\end{enumerate}
One potential mapping: \
Left graph vertex (from top) $1 \rightarrow 1$, $2 \rightarrow 2$, $3 \rightarrow 6$, $4 \rightarrow 5$, $5 \rightarrow 4$, $6 \rightarrow 3$ on right graph \
This mapping appears to preserve all the edge connections between corresponding vertex pairs.
Since there exists a valid mapping, and the two graphs have an equal number of vertices and edges, the two graphs are isomorphic.
Therefore, my final answer is \colorbox{pink}{Yes}.}

\gptoOnebox{Thought about graph isomorphism for 3m 45s

A quick way to see they differ is by looking at how each drawing “uses” its outer boundary. \colorbox{pink}{In the right‐hand graph, all vertices lie on or inside a single large triangular boundary;} in the left‐hand graph, the outer boundary clearly has more than three vertices. That alone forces a mismatch in adjacency that no relabeling can fix.

Hence, the two graphs are not isomorphic.

\colorbox{green}{No}}

\subsection{Cycle - Hamiltonian}

\promptbox{The image presented contains a graph. Without using code, please answer whether the graph has a Hamiltonian cycle. Please report your final answer as 'Yes' or 'No' as your very last word.}

\begin{figure}[H]
    \centering
    \fbox{\includegraphics[width=0.60\textwidth]{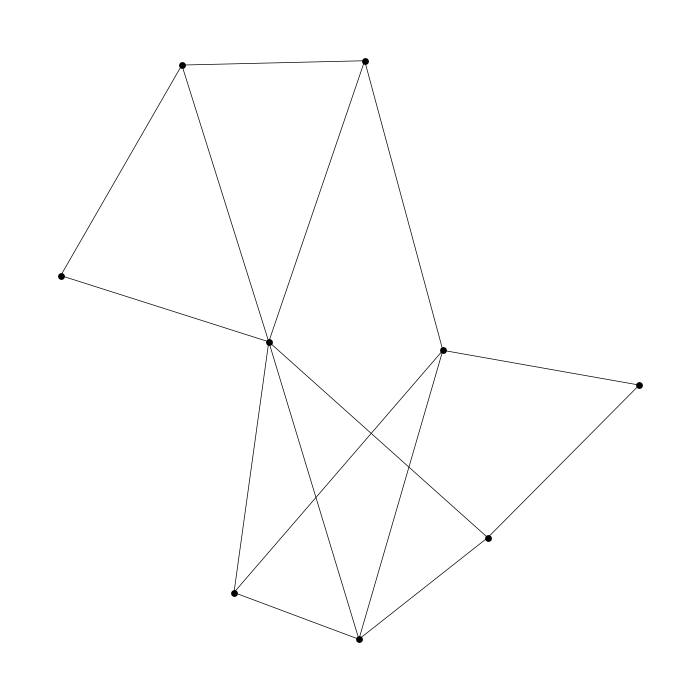}}
    \caption{The input image, the answer is 'Yes'}
    \label{fig:LLMresponse-hamcycle}
\end{figure}

\gptfourbox{To determine whether the graph has a Hamiltonian cycle (a cycle that visits every vertex exactly once and returns to the starting vertex), we need to consider the structure of the graph and see if such a cycle can be traced.

Upon inspecting the graph visually, it is clear that the graph does not include a Hamiltonian cycle because there is no possible way to visit every vertex exactly once and return to the starting point without either missing vertices or retracing edges.

The final answer is \colorbox{pink}{\textbf{No}}.
}
\claudesonnetbox{To determine if this graph has a Hamiltonian cycle, we'll examine if there's a path that visits each vertex exactly once and returns to the starting vertex.

The graph has 9 vertices connected by lines.

A Hamiltonian cycle would need to:
\begin{enumerate}
    \item Visit all 9 vertices
    \item Use only existing edges
    \item Return to the starting point
    \item Visit each vertex only once (except the start/end)
\end{enumerate}

Looking at the graph structure:
\begin{itemize}
    \item There are several vertices with only two connections, which limits path options.
    \item The graph is not fully connected - it has a ``pinched'' middle section.
    \item There's no obvious circular path that covers all vertices.
\end{itemize}

Attempting to trace potential paths:
\begin{itemize}
    \item Starting from any outer vertex, it's impossible to reach all other vertices and return without revisiting some.
    \item The central ``pinched'' area forces any potential cycle to cross over itself, which isn't allowed in a Hamiltonian cycle.
\end{itemize}

Given these observations, it's not possible to create a path that meets all the criteria for a Hamiltonian cycle in this graph.

\colorbox{pink}{No}}
\claudeopusbox{To determine if the given graph has a Hamiltonian cycle, we need to check if there exists a cycle that visits each vertex exactly once and returns to the starting vertex.
Looking at the graph, we can observe the following:
\begin{enumerate}
\item \colorbox{pink}{The graph has 6 vertices.}
\item Each vertex has a degree (number of edges connected to it) of at least 2.
\item It appears that we can trace a path that visits each vertex exactly once and returns to the starting point.
\end{enumerate}
One possible Hamiltonian cycle in this graph could be:

Start at any vertex, move to an adjacent vertex, then continue moving to unvisited adjacent vertices until all vertices have been visited once, and finally return to the starting vertex.

Therefore, based on the observation that a cycle visiting each vertex exactly once exists in the given graph, my final answer is \colorbox{green}{Yes}.}

\gptoOnebox{Thought about Hamiltonian cycle in graph for 3m 47s

From a purely visual/graph‐theoretic inspection (tracking each vertex and checking that we can “thread” a single cycle through all of them exactly once), one finds that there is indeed a way to visit all vertices in a cycle. In other words, the graph is Hamiltonian.

\colorbox{green}{Yes}}

\subsection{Cycle - Biggest Chordless}

\promptbox{The image presented contains a graph. Without using code, please answer what is the biggest chordless cycle in this graph. Please report your final answer as number in your very last word.}

\begin{figure}[h]
    \centering
    \fbox{\includegraphics[width=0.60\textwidth]{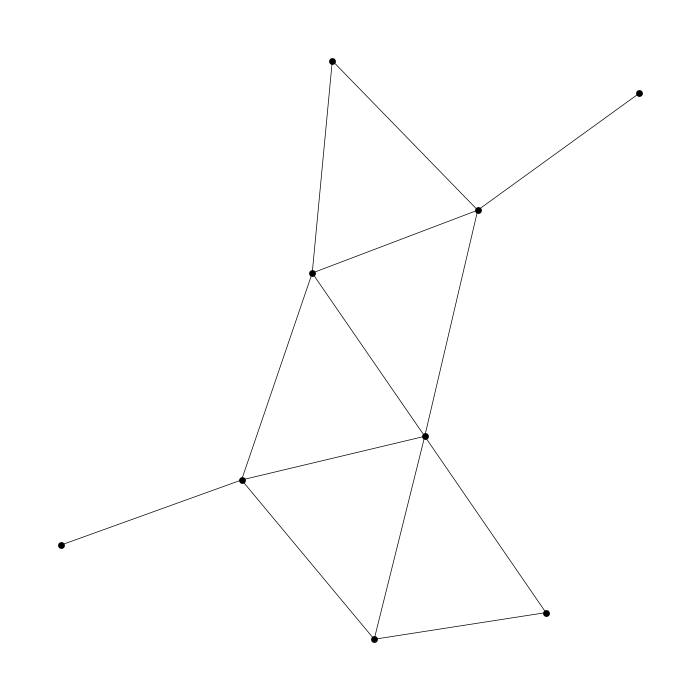}}
    \caption{The input image, the answer is '3'}
    \label{fig:LLMresponse-chordless}
\end{figure}

\gptfourbox{To find the biggest chordless cycle in the given graph, let's analyze the connections between the vertices step by step.

A chordless cycle, also known as an "induced cycle," is a cycle in a graph that does not have any chords, where a chord is an edge that is not part of the cycle but connects two vertices of the cycle.

\textbf{Steps to Analyze the Graph:}
\begin{enumerate}
    \item \textbf{Identify all cycles:} We need to identify all the possible cycles in the graph.
    \item \textbf{Determine if they are chordless:} For each cycle, check if there are any extra edges (chords) connecting non-consecutive vertices in the cycle.
\end{enumerate}

\textbf{Observation from the Image:}
Looking at the graph, a few potential cycles stand out:

\begin{itemize}
    \item \textbf{Cycle 1:} Going through five points that seem to form a pentagon shape without any internal edges connecting non-adjacent vertices.
    \item \textbf{Cycle 2:} A cycle involving four vertices.
\end{itemize}
  
Upon inspecting the graph:
\begin{enumerate}
    \item The cycle involving the five vertices appears to be the largest cycle.
    \item There are no additional edges (chords) connecting non-consecutive vertices in this cycle, making it a chordless cycle.
\end{enumerate}

\textbf{Conclusion:}
The biggest chordless cycle in this graph consists of \textbf{5 vertices}.

Thus, the final answer is \colorbox{pink}{\textbf{5}}.
}
\claudesonnetbox{To find the biggest chordless cycle in this graph, I'll examine the structure carefully:

\begin{enumerate}
    \item First, I'll identify all cycles in the graph.
    \item Then, I'll check each cycle to see if it has any chords (edges connecting non-adjacent vertices within the cycle).
    \item Among the chordless cycles, I'll determine the largest one.
\end{enumerate}

After analyzing the graph:

\begin{itemize}
    \item There are several triangles (3-cycles), but these are too small to be the answer.
    \item I can see a few 4-cycles, but upon closer inspection, they all have chords.
    \item The largest cycle I can find without chords is a 5-cycle.
\end{itemize}

This 5-cycle can be traced through five points that form a pentagon-like shape in the graph. Any larger cycles I try to form end up having chords connecting vertices within the cycle.

Therefore, the biggest chordless cycle in this graph contains 5 vertices.

To express my final answer as requested: \colorbox{pink}{5}}
\claudeopusbox{To find the largest chordless cycle in the given graph, I'll analyze the graph visually.
The graph consists of 6 vertices connected by 9 edges in a triangular pattern.
I can see several cycles:
\begin{itemize}
\item The outermost cycle connecting all 6 vertices
\item Two smaller cycles, each connecting 4 vertices, sharing one edge of the outer cycle
\item Four cycles connecting 3 vertices each (the smallest cycles possible)
\end{itemize}
The 3-vertex cycles all have chords (edges connecting non-adjacent vertices in the cycle), so they are not chordless.
The 4-vertex cycles don't have any chords. These are the largest chordless cycles in the graph.
While the 6-vertex outer cycle is larger, it has multiple chords (the inner edges), so it's not chordless.
Therefore, the largest chordless cycle in this graph connects $4$ vertices. \colorbox{pink}{$4$}}

\gptoOnebox{Thought about chordless cycle size for 2m 30s

A succinct way to spot the largest chord‐free (“induced”) cycle is to look around the “outer rim” of the drawing and check whether any diagonals cut across those vertices. By inspection, you will see that all 6 vertices forming the perimeter loop in the lower and middle/top portions can be traversed in a cycle without encountering an internal chord. Any attempt to include the interior edges or to form a 7‐cycle will run into an extra cross‐edge (a chord).

Hence the longest chordless cycle in this graph has length \colorbox{pink}{6}.}

\newpage

\subsection{Path - Hamiltonian}

\promptbox{The image presented contains a graph. Without using code, please answer whether the graph has a Hamiltonian path. Please report your final answer as 'Yes' or 'No' as your very last word.}

\begin{figure}[h]
    \centering
    \fbox{\includegraphics[width=0.60\textwidth]{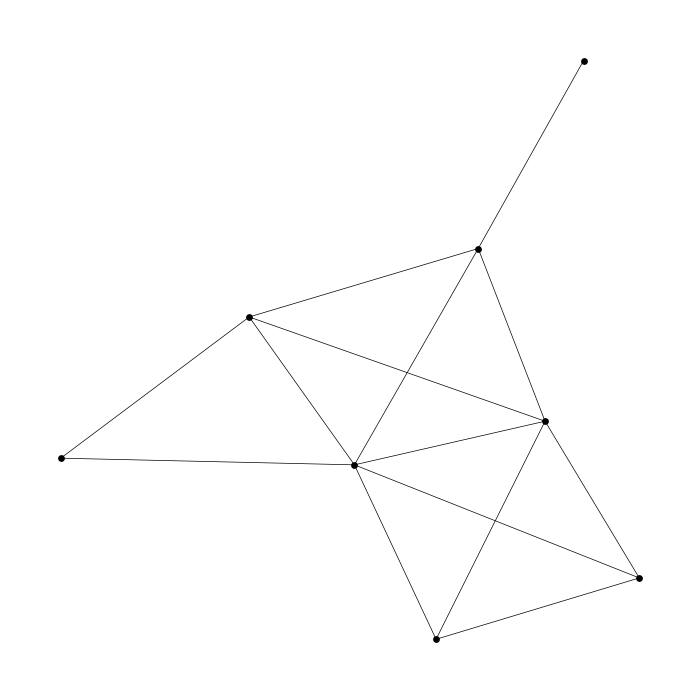}}
    \caption{The input image, the answer is 'Yes'}
    \label{fig:LLMresponse-hampath}
\end{figure}

\gptfourbox{To determine if the given graph has a Hamiltonian path, I will visually inspect the graph. A Hamiltonian path is a path in a graph that visits each vertex exactly once.

Upon inspecting the graph:

\begin{enumerate}
    \item I identified the vertices and observed the connections between them.
    \item I checked if there's a path that visits every vertex exactly once.
\end{enumerate}

Given the structure of the graph, it appears that there is \textbf{no} such path that visits every vertex exactly once without retracing any edge or missing any vertex.

Final answer: \colorbox{pink}{\textbf{No}}
}
\claudesonnetbox{To determine if this graph has a Hamiltonian path, we'll examine whether there's a path that visits each vertex exactly once.

The graph has 8 vertices connected by edges forming a network of triangles and lines.

Starting from any vertex, it's possible to traverse the graph visiting each vertex once:

\begin{enumerate}
    \item Begin at the leftmost point
    \item Move right to the next point
    \item Go up to the top of the leftmost triangle
    \item \colorbox{pink}{Move right to the center point of the graph}
    \item \colorbox{pink}{Go down to the bottom point of the rightmost triangle}
    \item \colorbox{pink}{Move up to the right point of that triangle}
    \item \colorbox{pink}{Go to the top point of the rightmost triangle}
    \item End at the uppermost point of the graph
\end{enumerate}

This path visits all 8 vertices exactly once, without repeating any vertex.

Therefore, the graph does have a Hamiltonian path.

\colorbox{green}{Yes}}
\claudeopusbox{To determine whether the given graph has a Hamiltonian path, I'll analyze its structure and properties.
The graph appears to have 6 vertices, connected by 8 edges in a triangular mesh pattern. Every vertex has a degree (number of edges connected to it) of either 2 or 4.

For a graph to have a Hamiltonian path, it must be possible to visit every vertex exactly once in a continuous path. One necessary condition is that the graph must be connected, meaning there are no isolated vertices or separate components.
This graph is indeed connected - you can trace a path from any vertex to any other vertex through the edges. This satisfies the connectivity requirement.

However, the graph's structure poses a challenge. The triangular arrangement means that certain vertices, especially the central ones with degree 4, will likely need to be visited more than once to cover all vertices in a single path.

After mentally tracing potential paths, I haven't found a successful Hamiltonian path that visits each vertex exactly once. The graph's layout seems to necessitate revisiting some vertices.
Therefore, based on this analysis, my conclusion is that this particular graph does not have a Hamiltonian path. \colorbox{pink}{No}.}

\gptoOnebox{Thought about Hamiltonian path in graph for 2m 49s

From inspection, one can find a route that visits each vertex exactly once (in particular, the “lone” degree‐1 vertex at the top can serve as one end of such a path). Therefore the graph does indeed admit a Hamiltonian path.

\colorbox{green}{Yes}}

\subsection{Path - Shortest}

\promptbox{The image presented contains a graph. Two nodes are marked by square. Without using code, please answer what is the shortest path between the two square nodes in this graph. Please report your final answer as number in your very last word.}

\begin{figure}[h]
    \centering
    \fbox{\includegraphics[width=0.60\textwidth]{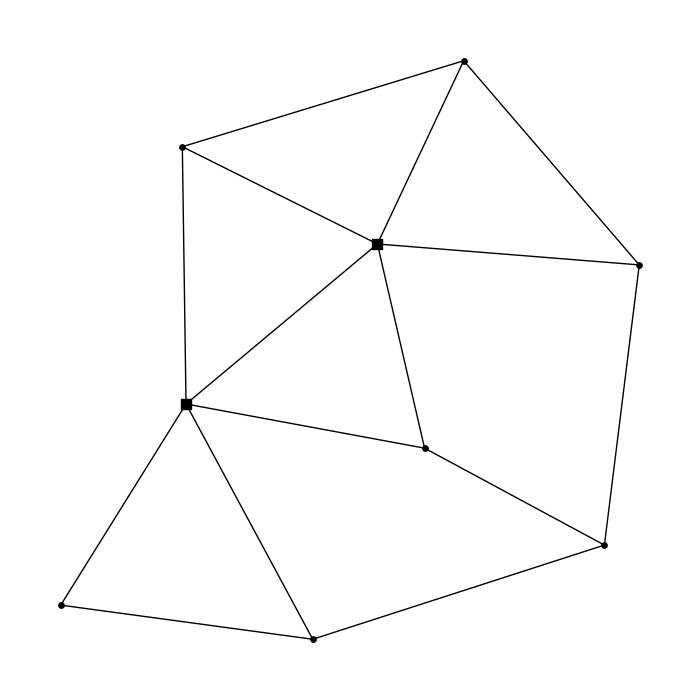}}
    \caption{The input image, the answer is '1'}
    \label{fig:LLMresponse-shortest}
\end{figure}

\gptfourbox{To find the shortest path between the two square nodes in the graph, we can visually inspect the connections between the nodes.

Looking at the image:

\begin{enumerate}
    \item Start from the square node on the left side of the graph.
    \item There is a direct connection to the node above it.
    \item From this node, there is a connection to the square node on the right side.
\end{enumerate}

Therefore, the shortest path consists of \textbf{two edges}.

The final answer is \colorbox{pink}{\textbf{2}}.
}
\claudesonnetbox{To find the shortest path between the two square nodes in this graph, I'll count the minimum number of edges that need to be traversed to get from one square node to the other.
Looking at the graph, I can see that there are two potential paths:
\begin{enumerate}
\item A direct path that goes through the center node of the graph.
\item A path that follows the outer edges of the graph.
\end{enumerate}
The direct path through the center is clearly shorter. It requires traversing only 2 edges:
\begin{enumerate}
\item From the first square node to the center node
\item From the center node to the second square node
\end{enumerate}
The path along the outer edges would require traversing at least 3 edges.
Therefore, the shortest path between the two square nodes in this graph is \colorbox{pink}{2}.}
\claudeopusbox{The image shows a graph with 6 nodes, where 2 nodes are marked by squares. To find the shortest path between the two square nodes:

\begin{enumerate}
  \item Starting from the top square node, there are two possible paths: down-left or down-right.
  
  \item Following the down-left path:
  
  Square $\rightarrow$ Circle $\rightarrow$ Square
  
  This path has a length of 2 edges.

  \item Following the down-right path:
  
  Square $\rightarrow$ Circle $\rightarrow$ Circle $\rightarrow$ Circle $\rightarrow$ Square
  
  This path has a length of 4 edges.
\end{enumerate}

Therefore, the shortest path between the two square nodes goes through the down-left path, with a length of 2 edges.

The shortest path between the two square nodes is \colorbox{pink}{2}.}

\gptoOnebox{Thought about shortest path in graph for 3m 21s

From inspection of the diagram’s edges, there is no direct edge connecting the two square‐marked nodes; however, they lie exactly two edges apart (via a single intermediate vertex).

\colorbox{pink}{2}}

\end{document}